\documentclass[twocolumn]{article}
\usepackage[twocolumn,textwidth=18cm,columnsep=1cm]{geometry}

\usepackage[dvipsnames]{xcolor}
\usepackage{amsmath,amssymb,amsfonts}
\usepackage{textcomp}
\usepackage{wrapfig}

\usepackage{cite}
\usepackage[pdftex]{graphicx}

\usepackage{algorithm}
\usepackage[noend]{algpseudocode}

\makeatletter
\def\BState{\State\hskip-\ALG@thistlm}
\makeatother

\usepackage{array}
\usepackage{multirow}
\usepackage{caption}
\usepackage{subcaption}
\usepackage{booktabs}
\newcommand{\coordinateSystem}{C}
\newcommand{\coordinateDevice}{\coordinateSystem_D}
\newcommand{\coordinateMagnet}{\coordinateSystem_M}
\newcommand{\magneticFluxDensity}{\mathbf{B}^{M}}

\newcommand{\figref}[1]{Fig. \ref{#1}}
\newcommand{\tabref}[1]{Tab. \ref{#1}}
\renewcommand{\eqref}[1]{Eq. \ref{#1}}
\newcommand{\secref}[1]{Sec. \ref{#1}}
\renewcommand{\algref}[1]{Alg. \ref{#1}}

\def\BibTeX{{\rm B\kern-.05em{\sc i\kern-.025em b}\kern-.08em
    T\kern-.1667em\lower.7ex\hbox{E}\kern-.125emX}}

\definecolor{abstractbg}{rgb}{0.89804,0.94510,0.83137}
\begin{document}

\title{Magnetic Tracking via Deep Learning}
\title{Magnetic Tracking via Deep Learning and Synthetic Data}
\title{Magnetic Tracking via Deep Learning and Finite Element Methods}
\title{Low-Latency 5-DoF Tracking of Non-Conventional Magnetic Markers via Deep Learning and Finite Element Methods}
\title{Low-Latency 5-DoF Tracking of Non-Spherical Magnetic Markers via Neural Networks and Finite Element Methods}
\title{5-DoF Tracking of Non-Spherical Magnetic Markers via Neural Networks}
\title{Neural Networks for Non-Spherical Permanent Magnet Localization}
\title{Single-step Tracking of Cylindrical and Spherical Magnets via Supervised Learning}
\title{Single-step Tracking of Axis-Symmetric Magnets via Supervised Learning}
\title{Robust Real-Time Tracking of Axis-Symmetric Magnets via Neural Networks}
\title{Robust 5-DoF Axis-Symmetric Magnetic Marker Localization Using Supervised Learning}
\title{Robust 5-DoF Magnetic Marker Tracking Using Synthetic Data and Supervised Learning}
\title{Utilizing Synthetic Data in Supervised Learning for Robust 5-DoF Magnetic Marker Localization}
%
%
%

\author{Mengfan~Wu\textsuperscript{*},
        Thomas~Langerak\textsuperscript{*},
        Otmar~Hilliges,
        Juan~Zarate
\thanks{\textsuperscript{*} Equal Contribution}
\thanks{This work was supported in part by grants from ERC (OPTINT StG2016-717054).}
\thanks{This work was conducted at the Department of Computer Science at ETH Z\"urich. Mengfan Wu is with TU Berlin and Huawei Research (email: mengfan.wu@campus.tu-berlin.de). All the other authors are with the Department of Computer Science at ETH Z\"urich. (email: first.last@inf.ethz.ch). All authors declare that they have no conflicts of interest.}}

\markboth{Transaction on Magnetics , VOL. XX, NO. XX, XXXX 2024}
{Wu \MakeLowercase{\textit{et al.}}: Utilizing Synthetic Data in Supervised Learning for Effective 5-DoF Magnetic Marker Localization}



\maketitle

\section{Abstract}

Tracking passive magnetic markers plays a vital role in advancing healthcare and robotics, offering the potential to significantly improve the precision and efficiency of systems. This technology is key to developing smarter, more responsive tools and devices, such as enhanced surgical instruments, precise diagnostic tools, and robots with improved environmental interaction capabilities.
However, traditionally, the tracking of magnetic markers is computationally expensive due to the requirement for iterative optimization procedures. Moreover, these methods depend on the magnetic dipole model for their optimization function, which can yield imprecise outcomes due to the model's significant inaccuracies when dealing with short distances between non-spherical magnet and sensor.
Our paper introduces a novel approach that leverages neural networks to bypass these limitations, directly inferring the marker's position and orientation to accurately determine the magnet's five degrees of freedom (5 DoF) in a single step without initial estimation. Although our method demands an extensive supervised training phase, we mitigate this by introducing a computationally more efficient method to generate synthetic, yet realistic data using Finite Element Methods simulations. The benefits of fast and accurate inference significantly outweigh the offline training preparation.
In our evaluation, we use different cylindrical magnets, tracked with a square array of 16 sensors. We perform the sensors' reading and position inference on a portable, neural networks-oriented single-board computer, ensuring a compact setup.
We benchmark our prototype against vision-based ground truth data, achieving a mean positional error of 4 mm and an orientation error of 8 degrees within a 0.2x0.2x0.15 m working volume. These results showcase our prototype's ability to balance accuracy and compactness effectively in tracking 5 DoF.

\section{Introduction}
Tracking the three-dimensional position and orientation of objects hidden from view is crucial for numerous applications. For instance, in Human-Computer Interaction (HCI), where the users' own body might occuled their hands, it involves decoding gesture input and delivering force feedback \cite{wearable_finger, handTrackingXIDIAN, finexus, magnetips, Omni, chen2021magx, langerak2020optimal, abler2021hedgehog}. In healthcare, tracking plays a vital role in guiding medical instruments such as capsules and catheters \cite{magnetic_catheter, magnet_capsule1, medical_instrument, magnet_capsule2, Luo2019}. Similarly, in robotics, tracking is used for tasks such as magnetomicrometry \cite{taylor2021magnetomicrometry} and tracking multi-arm robots \cite{song2021real}. 
Magnetic tracking enables the precise localization of objects hidden from view by utilizing magnetic markers. This is accomplished by measuring the magnetic fields around these markers with Hall sensors. The widespread interest in this technology is driven by its ability to permeate through various materials, including human tissues, without interference.

However, tracking magnetic markers is not a straightforward task. Existing research distinguishes between active and passive magnetic markers. Active markers emit a magnetic signal through an integrated power source like a battery, offering a higher signal-to-noise ratio and the advantage of frequency filtering. However, tracking active emitters requires either a wired connection to the tracked object or a battery-powered circuitry \cite{finexus, auraring}. Hashi et al. have proposed a solution to overcome this limitation with semi-passive markers \cite{wireless_marker}. While the tracked elements are passive, they respond with a resonant signal when excited by a nearby source. In contrast, passive magnetic localization is simpler, requiring only the target to be equipped with a permanent magnet, eliminating the need for tethered connections or active electronics on the moving elements \cite{Omni, cube_em_loc, multiMagnet}. Our work primarily focuses on passive magnetic localization due to its simplicity and robustness; however, future work could potentially extend the ideas presented here to other types of magnetic markers.

To track a magnetic marker we need to map magnetic signals, measured by one or more Hall sensors, to a marker's 5-degree-of-freedom (5DoF) position and orientation (rotation). The most straightforward approach involves directly interpolating the reading intensity within the sensor array. In \cite{gaussMarbles,gaussSense}, this approach is employed to track a stylus and a gaming object in a two-dimensional plane. In more complex configurations (e.g., \cite{utrack,finexus,magnetips}), multiple sensors with known spatial arrangements create an over-constrained system of equations, ensuring a unique solution for the magnet's state. These triangulation methods are fast but are not robust to sensor noise and limited to positional degrees of freedom.

A more robust approach, that also infers rotation, involves minimizing the difference between a theoretical magnetic field model, based on a magnetic model, and the sensor readings. To achieve this, several non-linear optimization algorithms have been proposed \cite{ChaoHu2005, ThanReview2012}. In \cite{multiMagnet}, the authors implemented an analytical computation of gradients to expedite the optimization process. In \cite{Yousefi2021}, their algorithm decouples the marker's orientation from its position, enhancing calculation speed and providing guarantees regarding global minima. Nonetheless, these iterative approaches come with significant challenges. Firstly, gradient-descent algorithms are computationally demanding, leading to a trade-off between tracking precision and frequency. Secondly, iterative non-convex optimization can be susceptible to the non-uniqueness of the solution, potentially converging to wrong local minima. These challenges make these methods highly dependent on their initialization.

Furthermore, these methods assume that the magnetic model, that predicts the theoretical magnetic field, is valid. All the mentioned methods employ the first term of the multi-pole series expansion derived from Maxwell's equations. The magnetic dipole approximation assumes all magnets to be spherical, providing the simplest explicit expression for the magnetic field concerning distance to the source. However, this approach has significant limitation. For non-spherical magnets, reliable results are only obtained when the magnets are far from the sensor \cite{Petruska2013}. Crucially, in gradient descent optimization process, a small magnetic field approximation error can lead to substantial positional discrepancies.

Recently, machine learning (ML) has emerged as a valuable approach to circumvent the computational burden of iterative methods. Machine learning, especially Neural Networks (NN), excels in approximating non-linearities through sequential multiply-addition operations. Additionally, it doesn't require initial estimates, often achieving reliable results with a single inference. While inference is rapid, creating a large and representative dataset for training to generalize well to unseen samples can be a challenging and time-consuming process. 

In \cite{Sebkhi2019, medical_instrument}, ML is utilized to predict the locations of tracked magnets using input from magnetic sensors. The data for training the neural network is collected in-vivo by placing magnets at various known locations and gathering sensor readings. The data collection process is lengthy and not scalable to new scenarios and markers. \cite{su2023amagposenet} employ a neural network to track a magnetic marker, though they train on synthetic data generated by the dipole model. While they eliminate the need for iterative gradient-based solutions, they do not address the limitations of the underlying model, retaining inaccuracies for non-spherical or closely located magnets. Recently, authors in \cite{auraring} and \cite{Sasaki2020} demonstrate the use of neural networks trained on synthetic data to track active markers. Although we share conceptual goals with these works, their results apply exclusively to active emitters and do not consistently outperform methods based on iterative optimization.

To address the data limitation challenge, we employ Finite Element Methods (FEM) to simulate the complete set of Maxwell's equations. FEM is computationally intensive, particularly when employing a fine mesh for accurate results. Therefore, it is unsuitable for real-time modeling. However, FEM can be leveraged to generate noise-free synthetic datasets for training machine learning models. In various other fields, the use of FEM-generated synthetic datasets has successfully trained neural networks, although not for magnetic tracking. Examples include mechanical deformations \cite{hyperelastic}, elastoplasticity \cite{solidMechanics}, material inspection \cite{stellDefect}, and nano-structures \cite{nanostructure}. To further reduce computation time while increasing dataset size, we focus on axis-symmetric markers. We enable this approach by introducing a coordinate transformation algorithm that capitalizes on our markers' symmetric properties, converting FEM-simulation results from 2D to 3D.

To evaluate the efficacy of our system, we evaluate our method in both simulation and experiments. In simulation we find that our method outperforms iterative methods. We also compare our FEM-based approach, to a method that is trained on magnetic-dipole generated data, where our method significantly outperforms the baseline. In real-world experiments we assess various cylindrical magnets with $5$ degrees of freedom using a lower-power portable Nvidia Jetson Nano, where sensing and tracking are performed in parallel. We achieve an averaged error of $4$ mm and an orientation error of less than $8$ degrees, with an interactive rate of $75$ Hz when using $8$ sensors.

In summary, our work contributes significantly to the magnetic tracking literature in five key ways: \emph{i)} We propose a supervised learning approach that enables real-time tracking of arbitrary axis-symmetric magnets using Neural Networks, overcoming the computational limitations of previous iterative solutions by approximating the inverse function of the magnetic field. \emph{ii)} To address the challenge of data limitation, we leverage Finite Element Methods (FEM) to simulate Maxwell's equations, generating noise-free synthetic datasets for training machine learning models. \emph{iii)} We introduce a coordinate transformation algorithm that capitalizes on our markers' symmetric properties, converting FEM-simulation results from 2D to 3D, enabling accurate high-resolution magnetic field data for training neural networks. \emph{iv)} Due to the computational efficiency of our approach, we enable low-power and portable applications. We demonstrate this by using a Nvidia Jetson Nano. \emph{v)} We demonstrate the validity of our approach with in-silico and real-world evaluations. These contributions collectively enable more accurate and efficient magnetic tracking solutions, offering the potential for the development of truly portable tracking devices.
\section{Method}
\begin{figure}[!t]
    \centering
    \includegraphics[width=0.7\linewidth]{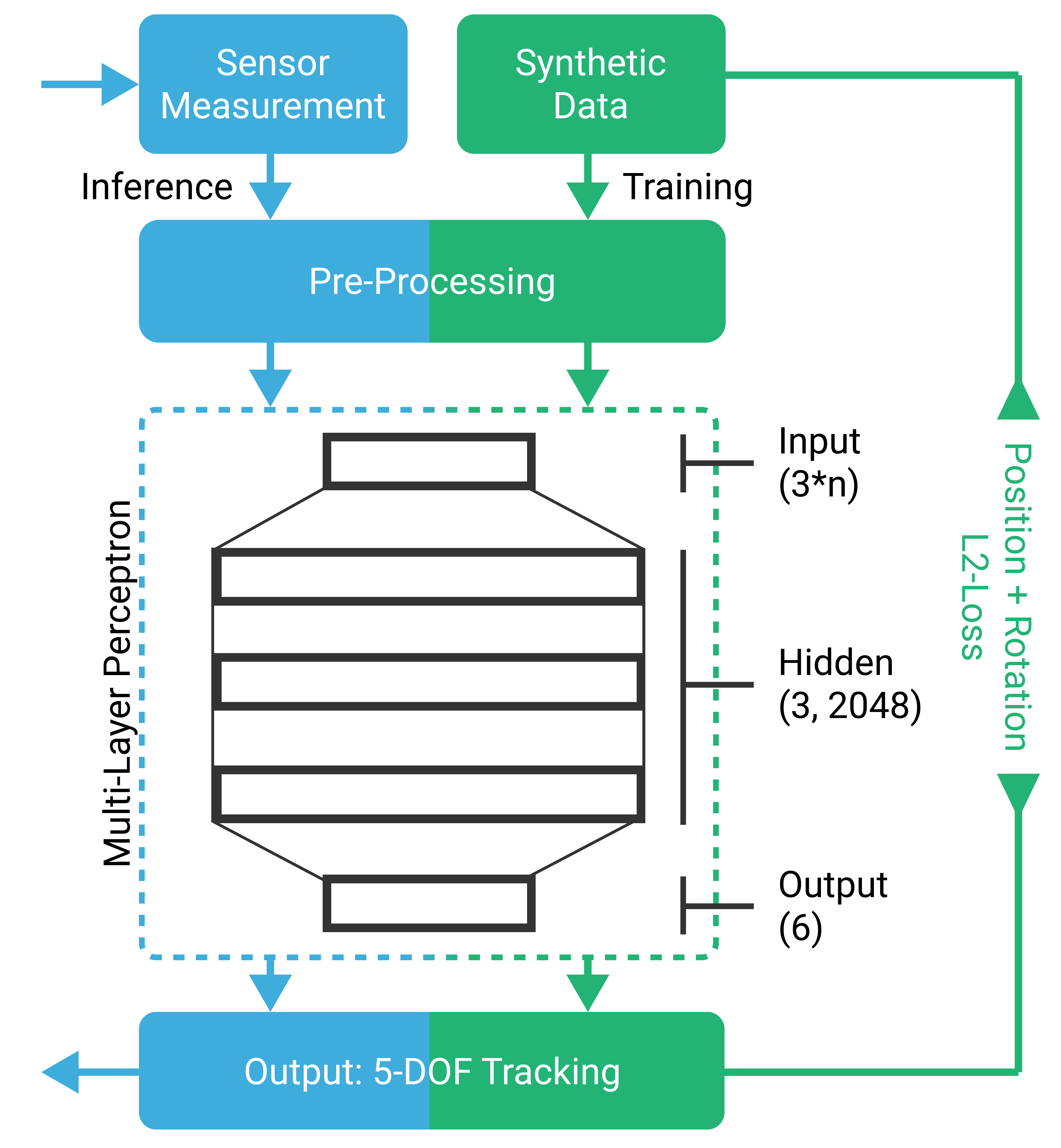}
      \caption[Multi-Layer Perceptron for Directly Predicting Magnet Positions]{Schematic  pipeline overview. We use a Multi-Layer Perceptron (MLP) to output the location and orientation of the magnet directly. During the training phase, the MLP outputs are compared with the ground truth for generating synthetic sensor readings. During the inference, the input to the MLP is the sensor data to track the position and orientation of the magnet.      \label{fig:mlp}}
\end{figure}

Our contribution revolves around a novel tracking method through supervised learning, specifically utilizing a Multi-Layer Perceptron (MLP). Our tracking pipeline consists of two distinct phases: the training phase involves comparing the MLP's output to the position and orientation used to generate synthetic sensor readings through Finite Element Methods (FEM), while during the inference phase the sensor readings are inputted into the MLP to determine the magnetic marker's position and orientation (as illustrated in \figref{fig:mlp}).

In this section, we provide a detailed explanation of our approach, starting with the creation of a high-resolution synthetic dataset (\secref{sec:fem_sim}). Subsequently, we detail the architecture and training process of our neural network (\secref{sec:track}). Finally, we outline the hardware setup (\secref{sec:hardware}).
\subsection{Synthetic Dataset}
\label{sec:fem_sim}
Our focus is on axisymmetric magnetic markers, encompassing any shape and size of magnets, provided they have rotational symmetry around their magnetization axis. This category encompasses cylinders, spheres, and toroids with arbitrary cross-sections, covering the most commonly used types of permanent magnets.

Instead of costly 3D simulations, we leverage the symmetry to conduct just one high-resolution 2D Finite Element Method (FEM) simulation for each magnet shape. From a single 2D FEM simulation, we generate synthetic sensor readings for any location and orientation of the magnet. We achieve the 3D volume by revolving this 2D cross-section around the magnet's principal axis. This approach of creating a synthetic dataset significantly enhances computational efficiency during training, striking an optimal balance between detailed simulation and minimal data storage requirements.

\begin{figure}
    \centering
    \includegraphics[width=0.8\columnwidth]{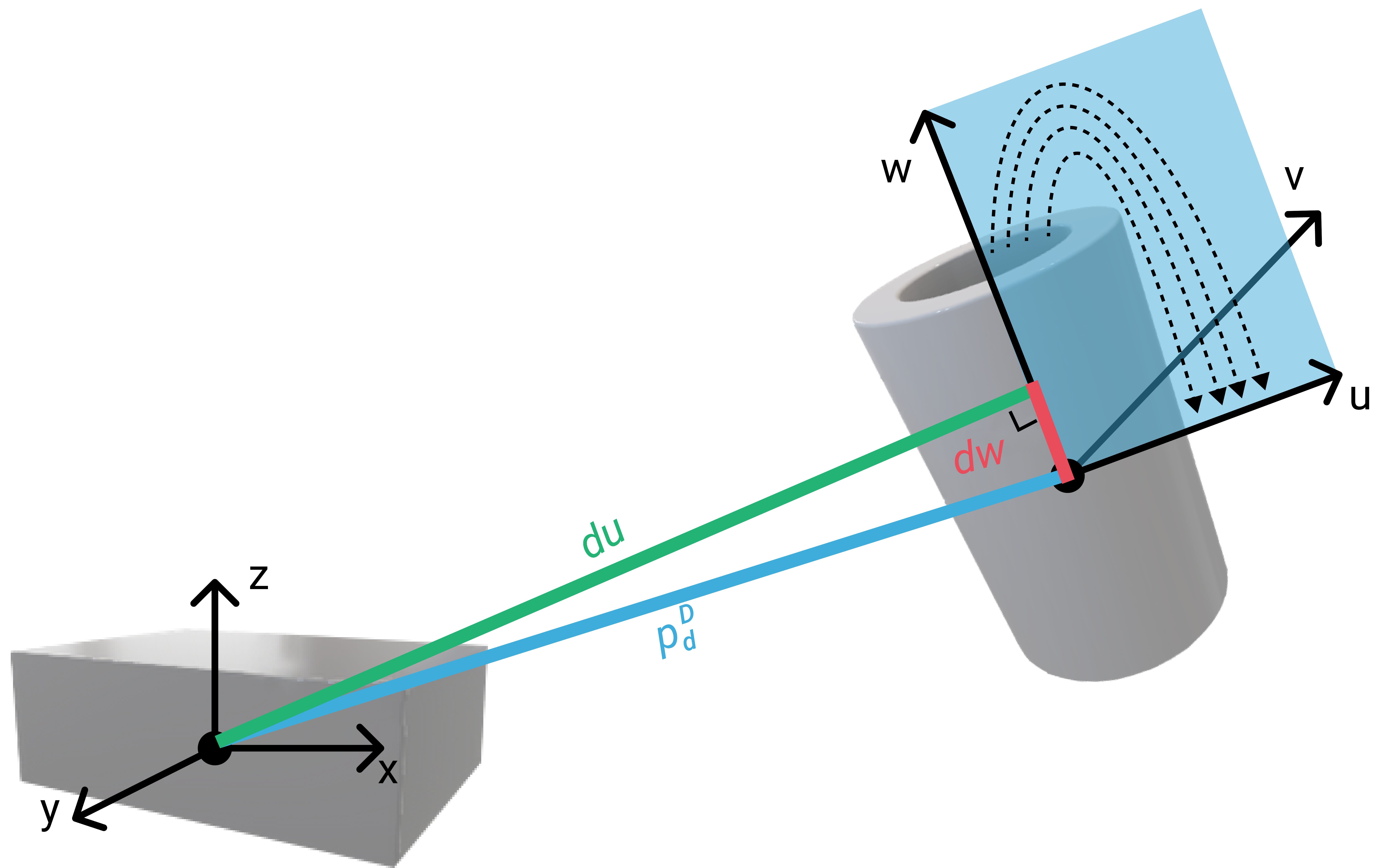}
    \caption{Coordinate system overview for \algref{algo:coor_trans}}
    \label{fig:coordinatealgo}
\end{figure}

We utilize COMSOL Multiphysics to acquire FEM data. The simulation centers on the magnet and due to the magnets symmetrical properties we constraint it to a single quadrant. We reconstruct the 3D magnetic B-field, using 2D FEM generated data and a sampled point. For generating training data and evaluating the neural network, we employ the transformation delineated in \algref{algo:coor_trans} (see \figref{fig:coordinatealgo}). This algorithm converts 2D FEM data into 3D synthetic sensor readings as follows. Initially, it requires the magnet's current position and the locations of each sensor, all within a fixed coordinate system, $\coordinateDevice$, centered at the device. Subsequently, we establish a coordinate system, $\coordinateMagnet$, centered on the magnet's current position and orientation. The three perpendicular axes of $\coordinateMagnet$ are along $\mathbf{u}$, $\mathbf{v}$ and $\mathbf{w}$ in $\coordinateDevice$. $\mathbf{u}$, $\mathbf{v}$ and $\mathbf{w}$ are then normalized to create coordinate transformation matrix $\mathbf{M}$. The algorithm then use $\mathbf{2D-FEM}$ to extract the magnetic flux density, $\magneticFluxDensity$ in $\coordinateMagnet$, with the sensor's coordinates in $\coordinateMagnet$ as $\langle du, 0, dw\rangle$, where the second coordinate is always 0 since we obtain FEM data in 2D. Ultimately, we transform $\magneticFluxDensity$ back into the original global coordinates, $\coordinateDevice$, to be used as features for training the MLP. All variables deployed in  \algref{algo:coor_trans} are explicated in \tabref{tab:var_trans}.

\begin{table*}[t]
    \centering
        \caption[Variables in Coordinate Transformation Algorithm]{Variables used in  \algref{algo:coor_trans}.\label{tab:var_trans}}
    \begin{tabular}{ll}
    \toprule
    \emph{Variables} & \emph{Description}\\
    \midrule
		$\mathbf{p}^{C}_i$ & Positional vector. For $^C$ and $_i$ see below.\\ \cmidrule{1-2}
		$C \in \{D, M\}$  & Coordinate system on device ($D$) or magnet ($M$)\\ \cmidrule{1-2}
		$i \in \{s, m\}$ & Positional vector of sensor ($S$) or magnet ($m)$\\ \cmidrule{1-2}
		$\mathbf{u, v, w}$ & {Axis in magnet's coordinate in $C_{D}$}\\ \cmidrule{1-2}
        $\mathbf{w}$ & \parbox[][][c]{5cm}{Magnetic moment direction in $C_{D}$, used as the third axis in $C_{M}$}\\ \cmidrule{1-2}
        $\mathbf{B}$ & Magnetic flux density\\ \cmidrule{1-2}
        $\mathbf{B}_n$ & Magnetic flux density detected at sensor $n$ \\ \cmidrule{1-2}
        \parbox[][][c]{2cm}{$B_n^d$,\\ $d \in \{x, y, z\}$} &\parbox[][][c]{5cm}{ $x$, $y$, or $z$-directional component of mangetic flux density detected at sensor $n$}\\
    \bottomrule
    \end{tabular}
    \end{table*}
    
\begin{algorithm}[H]
\caption{Synthethic dataset generation}\label{algo:coor_trans}
    \begin{algorithmic}[1]
    \BState \textbf{Input} $\mathbf{p}^{D}_s$
    \BState $\mathbf{p}_d^{D} \gets \mathbf{p}^{D}_m - \mathbf{p}^{D}_s$ \Comment Magnet-sensor vector
    \BState $\mathbf{v} \gets \mathbf{w} \times \mathbf{p}_d^{D}$
    \While {$\mathbf{v} = \mathbf{0}$} \Comment If $\mathbf{w}$ and $\mathbf{p}_d^{D}$ are (anti-)parallel
    \State random $\mathbf{q}$
    \State $\mathbf{v} \gets \mathbf{q} \times \mathbf{w}$
    \EndWhile
    \BState $\mathbf{u} \gets \mathbf{v} \times \mathbf{w}$
    \BState $dw \gets \mathbf{p}_d^{D} \cdot \frac{\mathbf{w}}{||\mathbf{w}||}$ \Comment the projection of $\mathbf{p}_d^{D}$ on $\mathbf{w}$
    \BState $du \gets \sqrt{|\mathbf{p}_d^{D}|^2 - dw^2}$ \Comment the projection of $\mathbf{p}_d^{D}$ on $\mathbf{u}$
    \BState $dv \gets 0 $   
    \BState $\mathbf{B}^{M} \gets \mathbf{2D-FEM}(du, dw)$
    \BState \parbox[][28pt][c]{232pt}{$\mathbf{M} \gets \big[\begin{smallmatrix}
      \frac{\mathbf{u}}{||\mathbf{u}||}^\top & \frac{\mathbf{v}}{||\mathbf{v}||}^\top & \frac{\mathbf{w}}{||\mathbf{w}||}^\top \end{smallmatrix}\big]^\top $ \\ \phantom{====} \Comment Coordinate transformation matrix}
    \BState $\mathbf{B}^{D} \gets \mathbf{B}^{M} \cdot \mathbf{M}$
    \end{algorithmic}
\end{algorithm}

\subsection{Tracking with Neural Networks}
\label{sec:track}

\subsubsection{Multi-Layer Perceptron}
\label{subsub:mlp}

We employ an MLP as our network architecture. The MLP takes in a $3n$-element vector as input, which contains the ($x, y, z$) magnetic flux densities $\mathbf{B}$ from the $n$ sensors. Its output is a 6-element tuple, comprising the magnet's position, $\mathbf{p} = [p_m^x, p_m^y, p_m^z]$, and orientation, $\mathbf{o} = [o_m^x, o_m^y, o_m^z]$. Therefore, we define the MLP as a non-linear mapping from the sensor readings to the tracking variables:

\begin{equation}
\mathcal{F}:\mathbb{R}^{3n}\rightarrow\mathbb{R}^6;\mathcal{F}(B_1^x, B_1^y, \ldots, B_n^z) =[\mathbf{p}, \mathbf{o}]
\end{equation}

It is important to note that, although we have a 5-Degree-of-Freedom (DoF) output, three in $\mathbf{p}$ and two in $\mathbf{o}$, we represent the orientation vector in Cartesian coordinates. This approach helps avoid numerical discontinuity, particularly when the azimuth angle transitions from $359^{\circ}$ to $0^{\circ}$. Our MLP architecture, depicted in \figref{fig:mlp}, consists of a 3-layer perceptron with 2048 units per layer, excluding the input and output layers. We use ReLU as the activation functions.

\subsubsection{Pre-processing}
Crucially, the dipole model (\eqref{eq:dipole}) indicates that the magnetic field decreases with the cube of the distance ($1/r^3$), where $r$ represents the distance between the source and the sensors, and $\mathbf{m}$ denotes the dipole moment. 
\begin{equation}
    \label{eq:dipole}
        {\mathbf{B}}({{\mathbf{p}}_m}) = \frac{{{\mu _0}}}{{4\pi }}\left[ {\frac{{3{\mathbf{r}}({\mathbf{m}} \cdot {\mathbf{r}})}}{{{r^5}}} - \frac{{\mathbf{m}}}{{{r^3}}}} \right],
\end{equation}

We have empirically determined that system training fails to converge when using magnetic readings directly as inputs. This issue may arise because the input values change dramatically, by several orders of magnitude, as the magnet moves from near a sensor (approximately $10^{-2}$ Tesla) to the boundary of the working volume (around $10^{-6}$ Tesla), consistent with the predictions of the dipole model where $\vert \mathbf{B}\vert \propto 1/r^3$.

\begin{figure}
    \centering
    \includegraphics[width=0.7\linewidth]{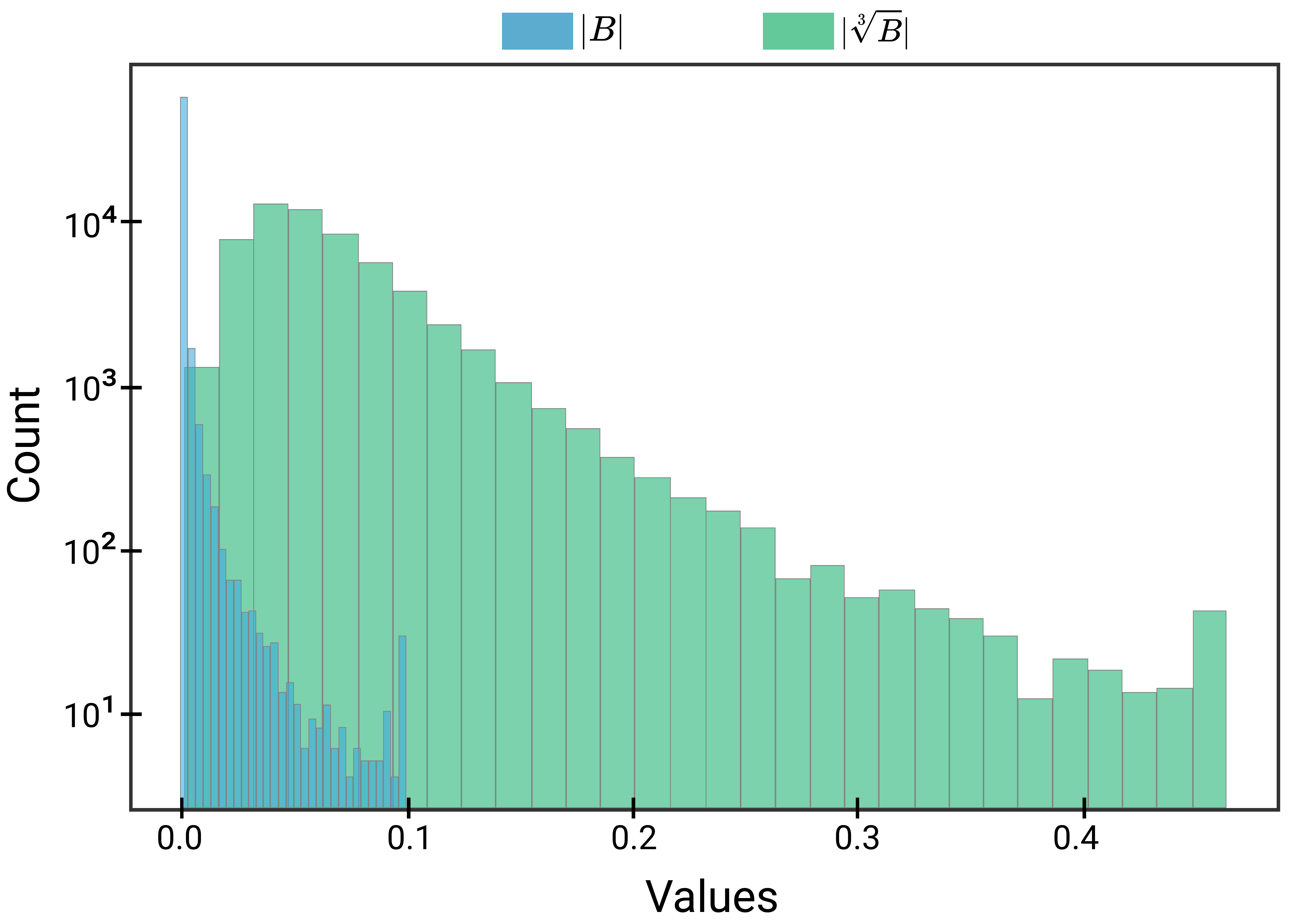}
    \caption[Histogram of value distribution before and after adding feature engineering function]{Histogram of input values before and after adding feature-engineering function.
      \label{fig:hist_cube}}
\end{figure}

To address this challenge we re-scale the input signals by taking their cubic root, $f(B) = \sqrt[3]{B} \propto 1/r$. The alteration in distribution is visible in \figref{fig:hist_cube}, where the values of $\sqrt[3]{B}$ is distributed on a wider range and more balanced. In general, an MLP benefits from having inputs that are more uniformly spread across the same order of magnitude. 

\subsubsection{Training}
\label{sec:training_nn}

We train our neural networks using randomly sampled data within a cubic volume measuring $0.2 \times 0.2 \times 0.15 m^3$, where the sensor array covers the bottom face. The magnets' orientations are sampled as points on a unit sphere and paired with their locations to serve as training labels, as outlined in \algref{algo:coor_trans}.

For our loss function, we employ a weighted sum of the positional and orientational differences:
\begin{equation} \label{eq:loss}
  \mathcal{L}= \|\mathbf{p}_{true} - \mathbf{p}_{pred}\|^2 + \eta \|\frac{\mathbf{o}_{true}}{\|\mathbf{o}_{true}\|} - \frac{\mathbf{o}_{pred}}{\|\mathbf{o}_{pred}\|}\|^2  
\end{equation}
\noindent Our goal is to minimize both the positional and orientational discrepancies between the predictions and ground truth. We introduce the weight $\eta$ to balance the differing scales of orientation and position error terms. After parameter tuning, we set  $\eta = 10^{-5}$ as the standard in all experiments.

\subsection{Implementation}
\label{sec:hardware}

\subsubsection{Software}
We implement the MLP in python with PyTorch. We use Adam \cite{kingma2017adam} with an initial learning rate $\gamma=10^{-4}$. 
We train for 40 epochs, and the learning rate decays by $0.98$ after each epoch in training.
We generate $10^6$ random points per epoch to train the model with a batch size of $256$.
Using \algref{algo:coor_trans} we generate the training data independently and identically distributed. Data generation and training of the MLP takes $\sim 1$ hour on a standard desktop.

\subsubsection{Hardware}
Our hardware's foundation consists of 16 triaxial magnetometers (MLX90393, Melexis). These sensors feature a linear range up to 0.05 Tesla, aligning well with the magnetic fields we anticipate from our test magnets at close range. These sensors can achieve a data output rate of 716.9 Hz. 

\begin{figure}[!t]
    \centering
    \begin{subfigure}[b]{0.7\columnwidth}
    \includegraphics[width=\textwidth]{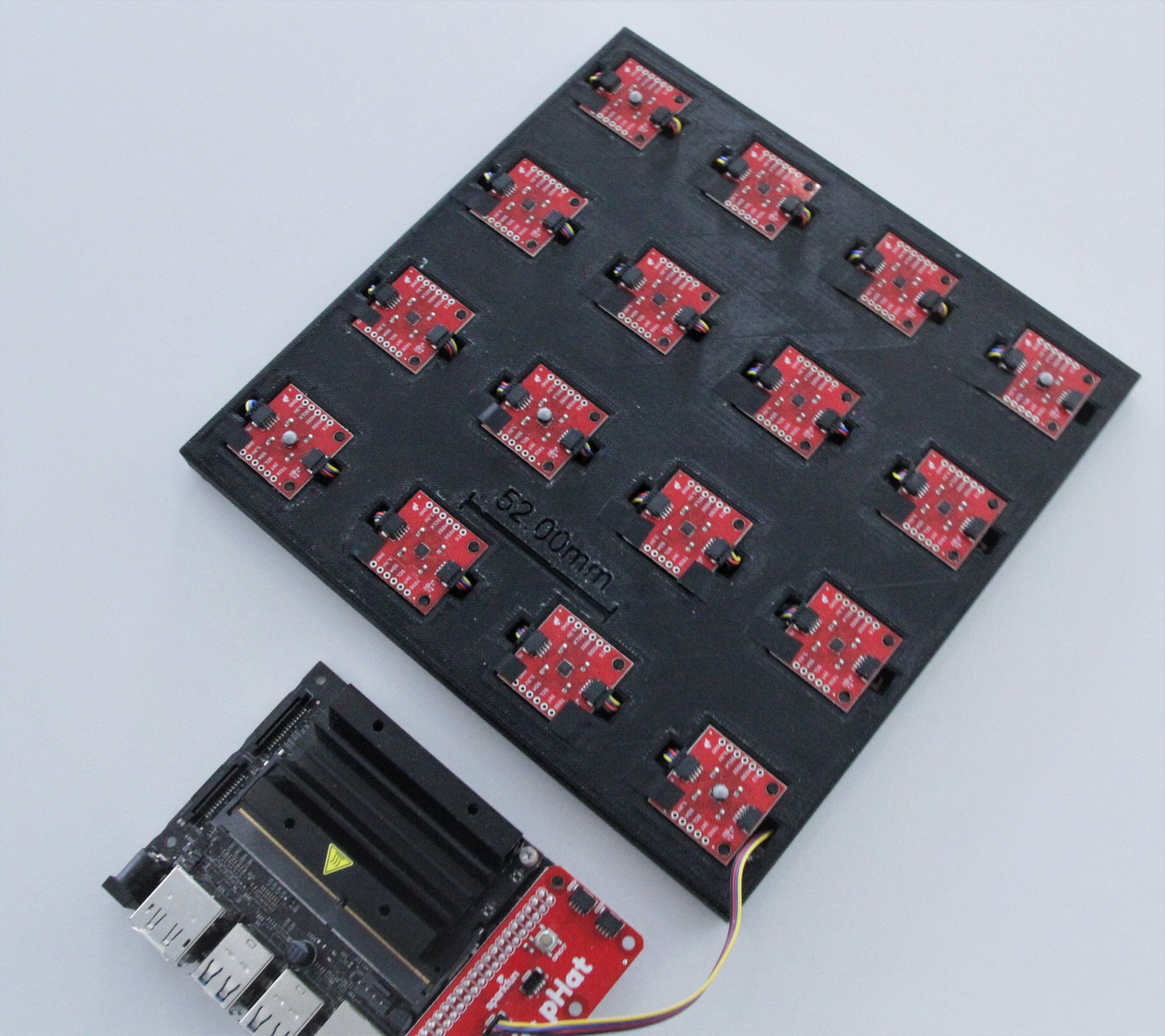}\llap{\makebox[\textwidth][l]{\raisebox{3.5cm}{\frame{\includegraphics[height=2.0cm]{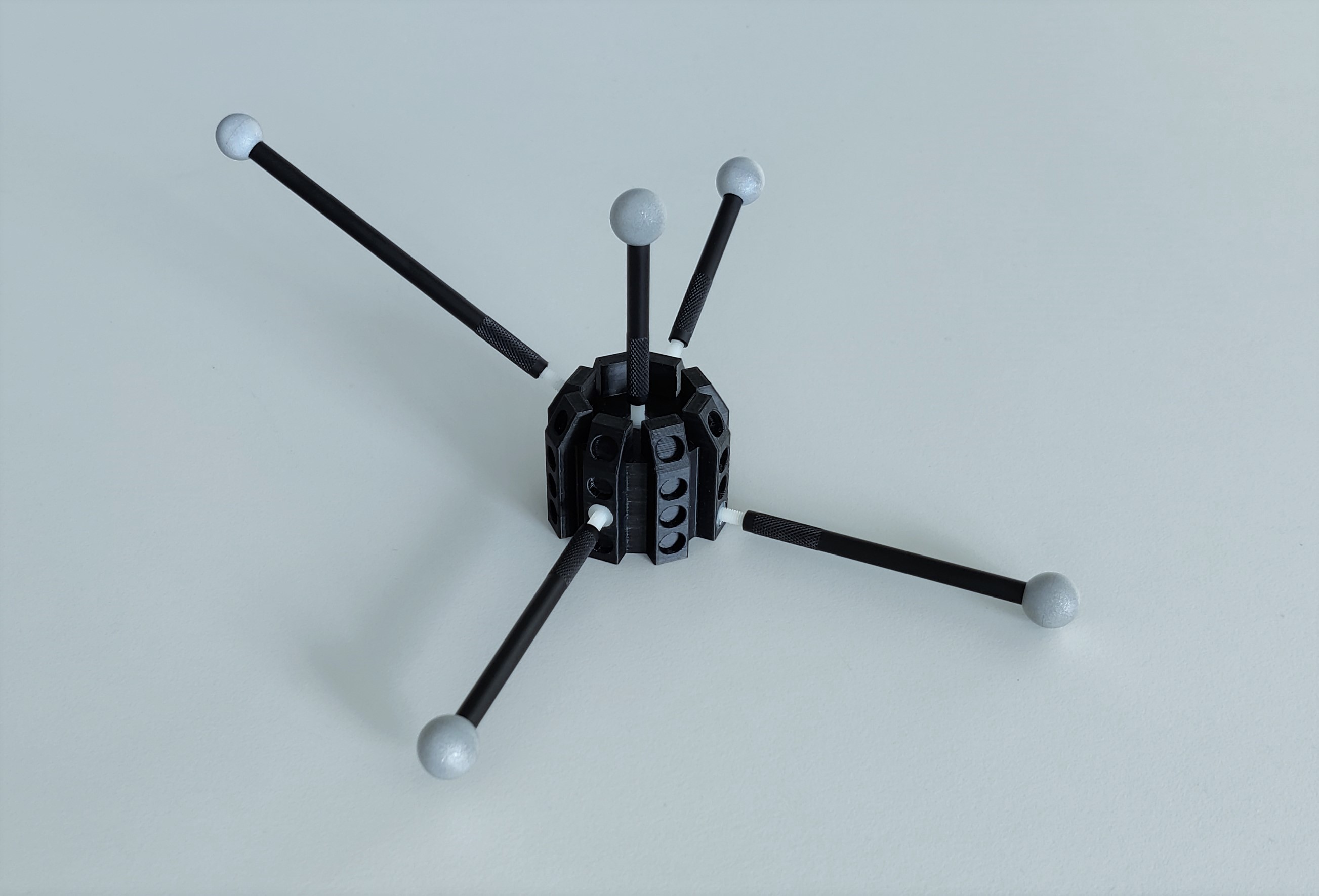}}}}}
    \end{subfigure}
    \caption[Placement of Sensors]{Hall sensor array in a 4x4 grid. The center-to-center distance is 52mm. The array is connected to a Jetson Nano for the complete inference pipeline. The inset shows a rigid-tree to collect Optitrack groundtruth. 
      \label{fig:sensor_loc}}
\end{figure}

Moreover, we calibrate our system using the ellipsoid-fitting method referenced in \cite{calibration}. The arrangement of the 16 sensors follows a 4 x 4 grid pattern, maintaining a 52 mm interval between the centers of adjacent sensors, as depicted in \figref{fig:sensor_loc}.

To maintain minimal instrumentation and facilitate portable applications, such as prosthetics, we have implemented the tracking inference on an AI-oriented single-board computer (Jetson Nano, NVIDIA). This device weighs only 138 grams and also supports direct sensor readings using the I2C communication protocol. The minimum time required to read 16 sensors sequentially is approximately 24 milliseconds.

\section{In-Silico Evaluation}
\label{sec:insilico}
In this section, we compare the performance of our neural network-based tracking method with that of an iterative, gradient-descent-based technique. This comparison utilizes simulated sensor readings, allowing us to control experimental variables such as the initialization of the iterative, optimization-based algorithm. We first introduce how we sample data points in order to evaluate both methods. Then we compare our method versus the iterative baseline in terms of accuracy as function of initialization, as well as computational time. Finally, we compare our method trained on FEM data versus trained on data generated with the dipole model.

\paragraph*{Sampling Method}
\label{sec:sample}
For the in-silico evaluation, we randomly sample magnet positions and orientations and compute the corresponding magnetic flux densities. We either use the magnetic dipole model \eqref{eq:dipole} to sample from or use FEM-generated data. On one hand, using dipole-model-generated data ensures that the characteristics of the synthetic signals matches the internal physical model used in the optimization method. On the other hand, the FEM-generated data is more realistic. We initialize the optimization method by offsetting the sampled position and orientation. We configure the iterative algorithm to stop upon reaching the maximum number of iterations. This process is similar to a real-world application, where we would initialize with the last known position and the newly measured magnetic flux densities, and we have a limited computational budget.

\subsection{Influence of initialization on the optimization methods}
\label{sec:synResults}

\begin{figure*}[t]
    \centering
        \begin{subfigure}[b]{0.4\textwidth}
        \includegraphics[width=\textwidth]{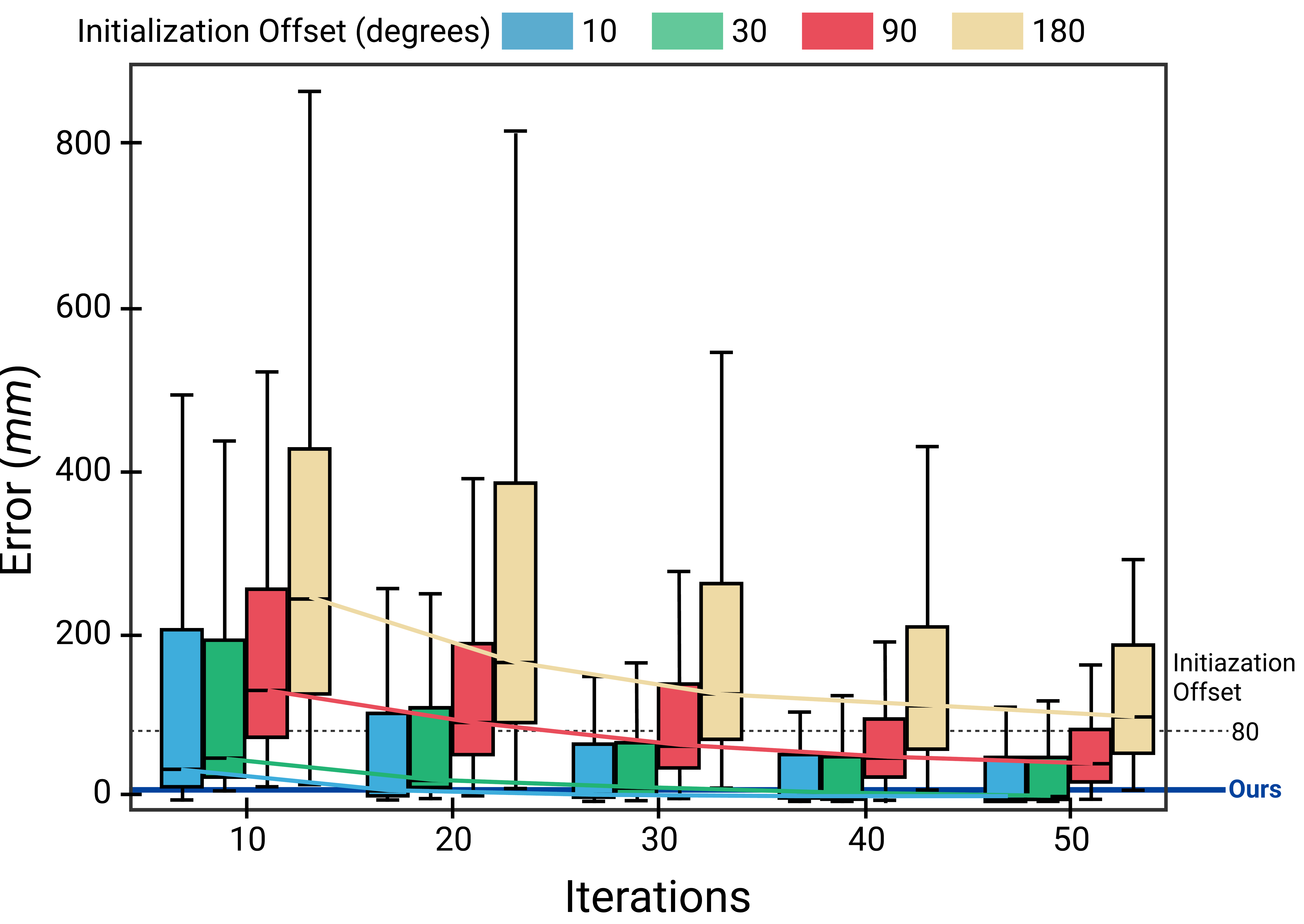}
        \caption{}
        \end{subfigure}
        \hspace{1cm}
        \begin{subfigure}[b]{0.4\textwidth}
        \includegraphics[width=\textwidth]{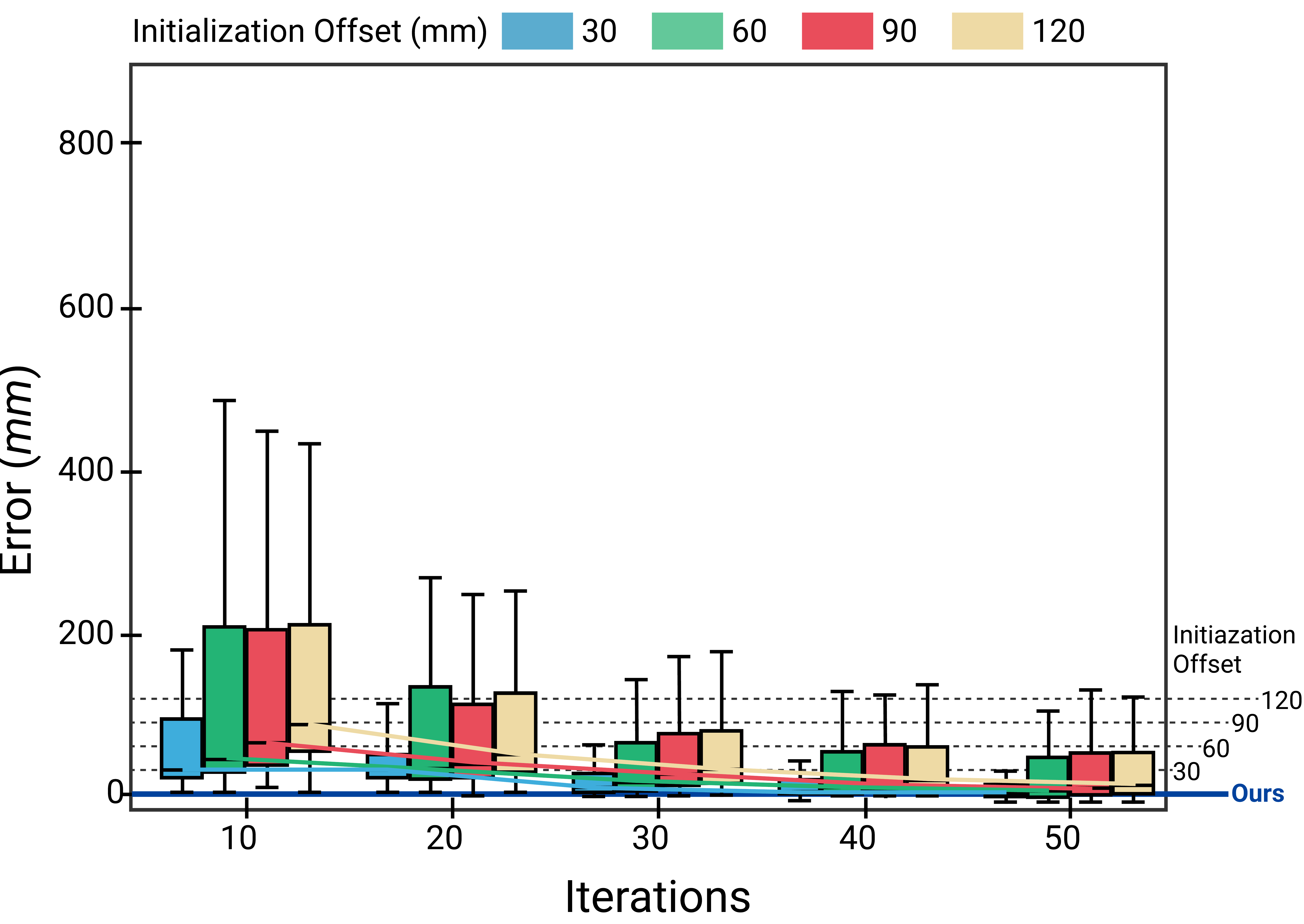}
        \caption{}
        \end{subfigure}

    \caption[]{
    Positional errors of the iterative method in simulation for different number of iterations.
         \textup{(a)} We vary the initial orientation mismatch, keeping a fixed distance of $80$ mm to the truth.
         \textup{(b)} We vary the initial positional mismatch, keeping a fixed orientation difference of $45^{\circ}$. 
         As our method does not rely on iterations or initialization, it is a single value. The standard deviation of our method is too small to show in the graph. 
      \label{fig:evo_angle_perturb}}
\end{figure*}

\begin{figure*}[t]
    \centering
        \begin{subfigure}[b]{0.4\textwidth}
        \includegraphics[width=\textwidth]{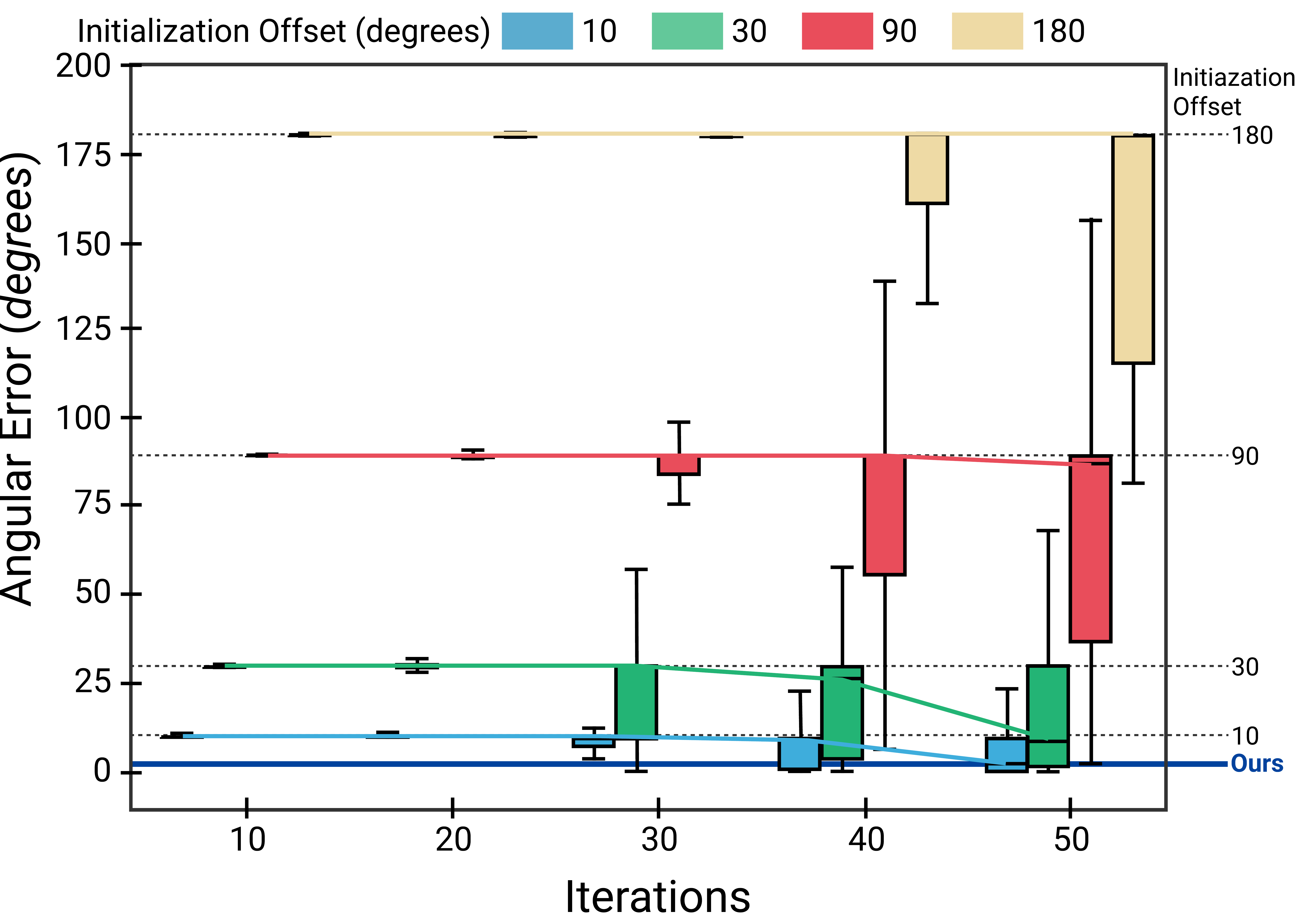}        
        \caption{}
        \end{subfigure}
        \hspace{1cm}
        \begin{subfigure}[b]{0.4\textwidth}
        \includegraphics[width=\textwidth]{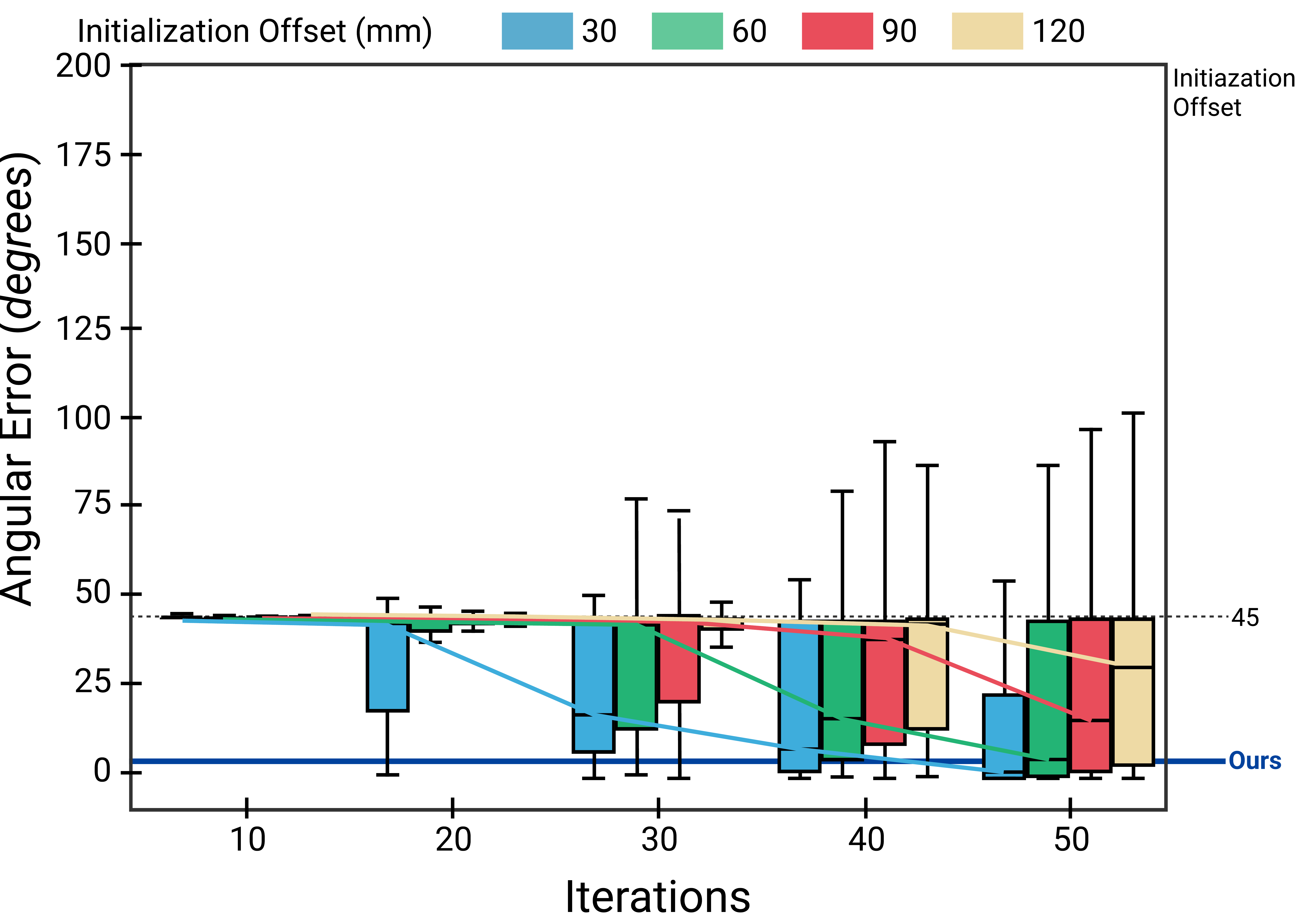}
        \caption{}
        \end{subfigure}

\caption[]{
    Angular errors of the iterative method in simulation for different number of iterations.
         \textup{(a)} We vary the initial orientation mismatch, keeping a fixed distance of $80$ mm to the truth.
         \textup{(b)} We vary the initial positional mismatch, keeping a fixed orientation difference of $45^{\circ}$. 
         As our method does not rely on iterations or initialization, it is a single value. The standard deviation of our method is too small to show in the graph. 
      \label{fig:evo_angular_error_perturb}}
\end{figure*}

In our simulation study, we conducted a comparison between our method and an optimization-based technique by analyzing $400$ data points, sampled as outlined above. For the optimization process, we utilized PyTorch, configuring the magnet's position and orientation as adjustable parameters. This setup leveraged automatic differentiation for efficient computation, following the approach detailed in \cite{auto_diff}. The optimization algorithm was L-BFGS, a quasi-Newton method. L-BFGS optimizes step size through a line search mechanism \cite{lbfgs}. For the optimization-based approach we employed the magnetic dipole model as our internal physical model, integrating it as our loss function in a manner consistent with \cite{multiMagnet}. Our results are presented in two parts: firstly, we assess the accuracy of our method compared to the optimization baseline for various initialization errors; secondly, we compare the tracking performance of our method trained on FEM data versus data generated by the dipole model.

\paragraph{Accuracy} The full results of our optimization method are detailed in \figref{fig:evo_angle_perturb}, showcasing the tracking errors relative to the maximum iteration count across different initialization of position (\figref{fig:evo_angle_perturb}.a) and orientation (\figref{fig:evo_angle_perturb}.b). Similar results are shown for the angular error (\figref{fig:evo_angular_error_perturb}) For a detailed analysis, we have also provided statistical analysis of the tracking outcomes after 50 iterations in Table \tabref{tab:sim_opt_stats}. Note that our MLP method is a single inference step, and does not require multiple iterations or an initialization. 

Since the non-converging cases in optimization method yield a long tail in error distribution, the average errors of optimization methods are not representative of the performance. Instead, we evaluate the results with the median value of errors. Our approach showed improved performance with a median positional error of \(e_p = 1.40mm\) and a median orientation error of \(e_\theta = 3.34^\circ\), significantly outperforming the baseline's median positional error of \(e_p = 26.80mm\) and orientation error of \(e_\theta = 29.96^\circ\) under nearly all test conditions. As expected, the accuracy of the optimization based approach improved with more iterations. This trend was also observed in relation to the quality of the initialization; more accurate initial conditions led to more precise optimization outcomes. However, our method was only surpassed in scenarios with the highest number of iterations (50) and the optimal initial conditions, specifically at 80 mm of initial positional error and \(10^\circ\) of initial orientation error, along with 30 mm of initial positional error and \(45^\circ\) of initial orientation error.

Decreasing the number of iterations led to a decrease in performance. With ten iterations, the median positional error rose above 200 mm in certain instances, especially when the initial orientation significantly deviated from the target. We observed that even with initial orientations closely aligned to the target (with an offset of \(45^\circ\)), the resulting positional errors on par with scenarios with a large initialization offset.

\begin{table*}[t]
    \caption[]{Accuracy results for the iterative method after 50 Iterations. For comparison, we include results obtained via our MLP tracking on the same evaluation data. Our method is not an iterative approach, and does not rely on an initialization}
    \label{tab:sim_opt_stats}
    \centering
    \begin{tabular}{>{\raggedleft}p{2cm}>{\raggedleft}p{2cm}|>{\raggedleft}p{2cm} >{\raggedleft}p{2cm} >{\raggedleft}p{2cm} >{\raggedleft\arraybackslash}p{1.3cm}}
    \hline
    Initial $e_p$ in mm &
    Initial $e_{\theta}$ in degrees &
    $\ \mathtt{median}$ $e_p$ in mm &
    $\mathtt{median}$ $e_{\theta}$ in degrees &
    $\mathtt{3^{rd} quartile}$ $e_p$ in mm &
    $\mathtt{3^{rd} quartile}$ $e_{\theta}$ in degrees \\
    \hline
    \multicolumn{6}{l}{}\\
    \multicolumn{2}{l}{\textit{\textbf{Iter. Optimization}}}\\
    \hline
    \hline
    80& 10& 0.65& 1.17& 11.90& 9.98\\
    80& 30& 2.06& 4.49& 16.91& 29.76\\
    80& 90& 35.12& 86.00& 86.65& 90.00\\
    80& 180& 86.27& 179.75& 147.30& 179.99\\
    \hline
    30& 45& 0.53& 1.04& 8.50& 28.33\\
    60& 45& 2.96& 5.78& 22.38& 44.40\\
    90& 45& 5.82& 12.85& 33.88& 44.82\\
    120& 45& 10.37& 25.20& 65.97& 45.03\\
    \hline
    & all & 26.80& 29.96& 95.24& 91.34\\
    \hline
    \multicolumn{6}{l}{ }\\
    \multicolumn{2}{l}{\textit{\textbf{MLP (ours)}}}\\
    \hline
    \hline
    4 & sensors & 1.93& 4.96& 2.88& 7.78\\
    8 & sensors & 1.62& 4.28& 2.25& 6.54\\
    12 & sensors & 1.42& 3.38& 1.89& 5.11\\
    16 & sensors & \textbf{ 1.40}& \textbf{3.34}& \textbf{1.87}& \textbf{5.06}\\
    \hline
    \end{tabular}
\end{table*}

\paragraph{Running Time}
\label{sec:evaSynRuntime} 

Our Multilayer Perceptron (MLP) model delivers outcomes after a single inference step, which includes feature engineering, along with additions and multiplications within its hidden layers. On the other hand, optimization-based approaches necessitate the calculation of second-order gradients relative to the estimated positions at every iteration, proceeding until convergence is achieved or the maximum number of iterations achieved. We evaluated the speed differences between these two methodologies. To ensure a fair comparison, we executed both methods on the same hardware, specifically a laptop powered by an Intel i7-7500U CPU. 

\begin{figure}[!t]
    \centering
    \vspace{-.5cm}
    \includegraphics[width=0.7\columnwidth]{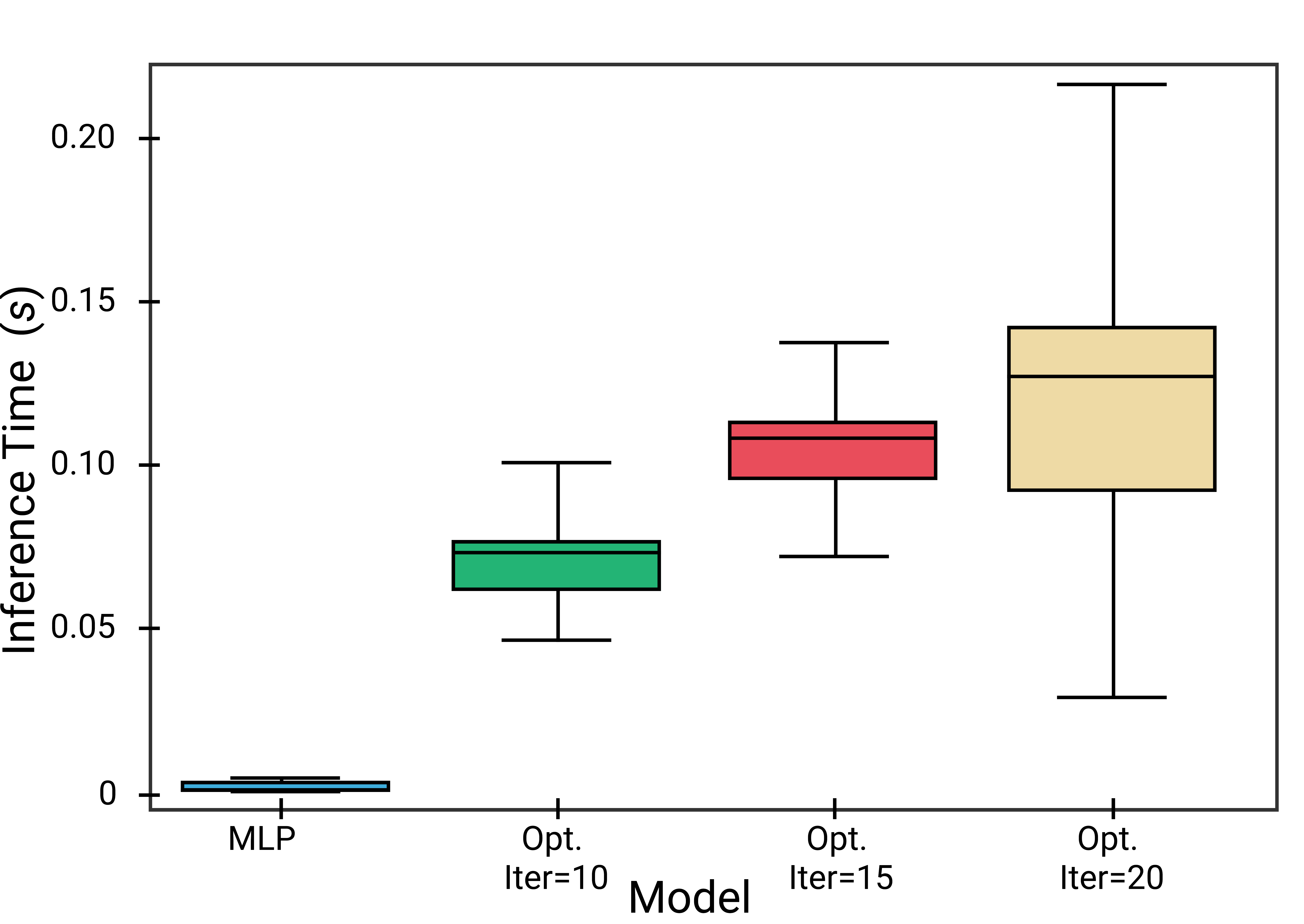}
    \caption[]{Inference time comparison between MLP and L-BFGS implementations for a different number of iterations. Iterative methods are strongly dependent on initialization and it takes between 20 and 50 iterations (i.e. $>$ 10 ms) to achieve results similar to those given by a single MLP inference (0.8 ms).
      \label{fig:running_time}}
\end{figure}

\figref{fig:running_time} illustrates the comparison between the total processing time of our data-driven MLP approach and the optimization-based method across various maximum iteration counts. A single inference operation for 5-degree-of-freedom (DoF) tracking using the MLP is completed in just 0.8 milliseconds (ms), this time frame includes the feature engineering stage. In contrast, the L-BFGS optimization process demands approximately 1 ms per iteration. As shown in \figref{fig:evo_angle_perturb}, achieving satisfactory outcomes with the optimization method often requires dozens of iterations, depending on how well the process is initialized. Consequently, the optimization approach can take tens of milliseconds to converge. Thus, it is evident that our method surpasses the iterative optimization baseline in terms of speed.

\subsection{Neural Networks trained with FEM vs Dipole Model}
\label{sec:synShapes}

Previous works train neural networks on data generated with the approximated analytical model \cite{su2023amagposenet,song2014magnetic}. In contrast, we propose using FEM to generate a dataset and take full advantage of the powerful representation of Neural Networks for non-linear systems. To ensure a fair comparison, we distinguish two scenarios for training the MLP: i) using FEM data, and ii) using the magnetic dipole model directly, which is similar with the model used in the optimization method.  

Our evaluation encompasses six different magnet shapes, similar to those in our practical experiments (\secref{sec:opt_eval}). Including, a variety of shapes from cylinders to disks, all magnetized along their principal axis, and we also assess performance on a spherical magnet, where the dipole model theoretically provides an exact solution. For each magnet shape, we generated a dataset consisting of $1000$ samples.

\begin{figure*}[t]
    \centering
    \begin{subfigure}[b]{0.4\textwidth}
        \includegraphics[width=\textwidth]{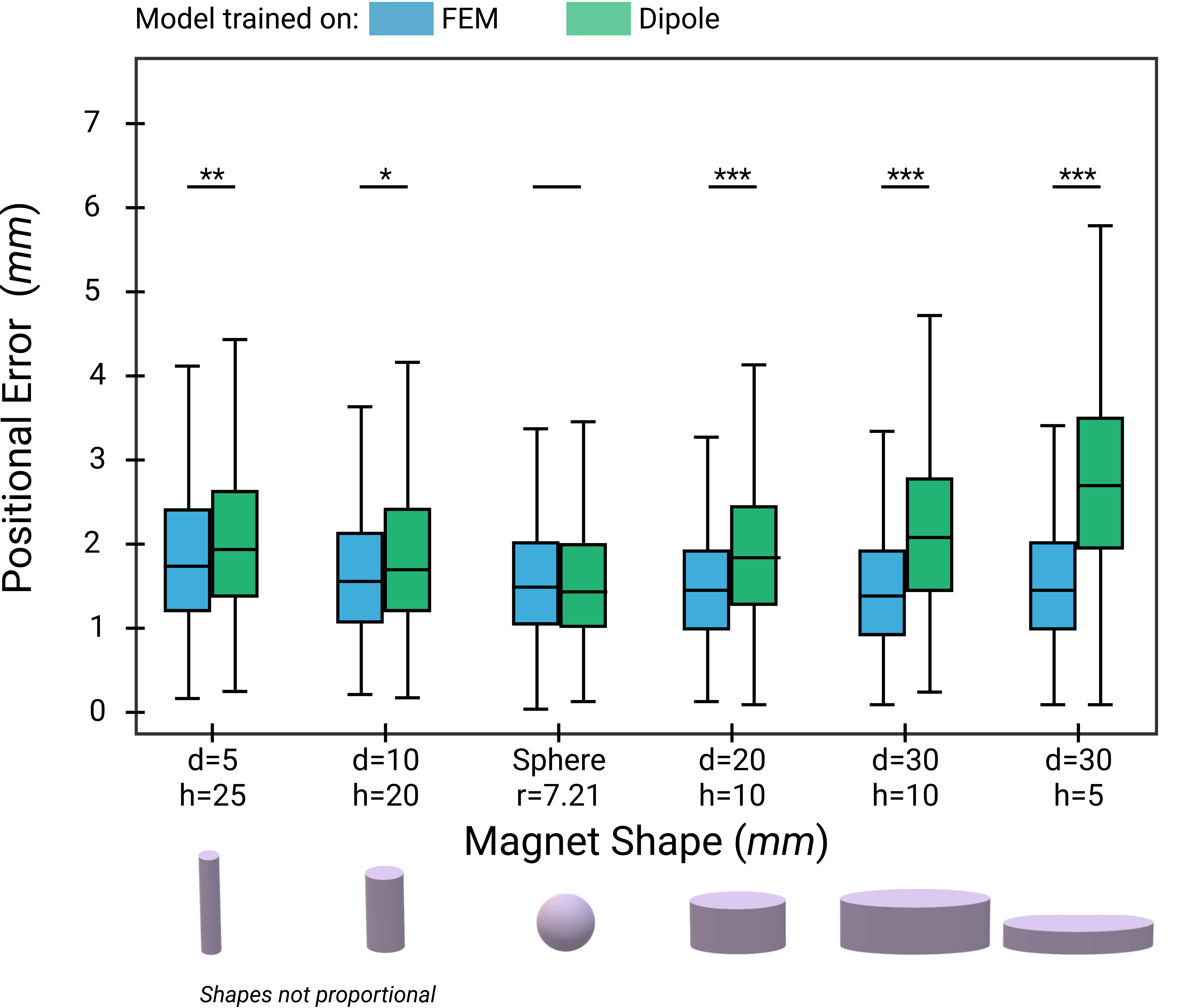}
        \caption{Positional errors}\label{fig:dipole_vs_mlp_nn}
    \end{subfigure}
    \hspace{1cm}
    \begin{subfigure}[b]{0.4\textwidth}
        \includegraphics[width=\textwidth]{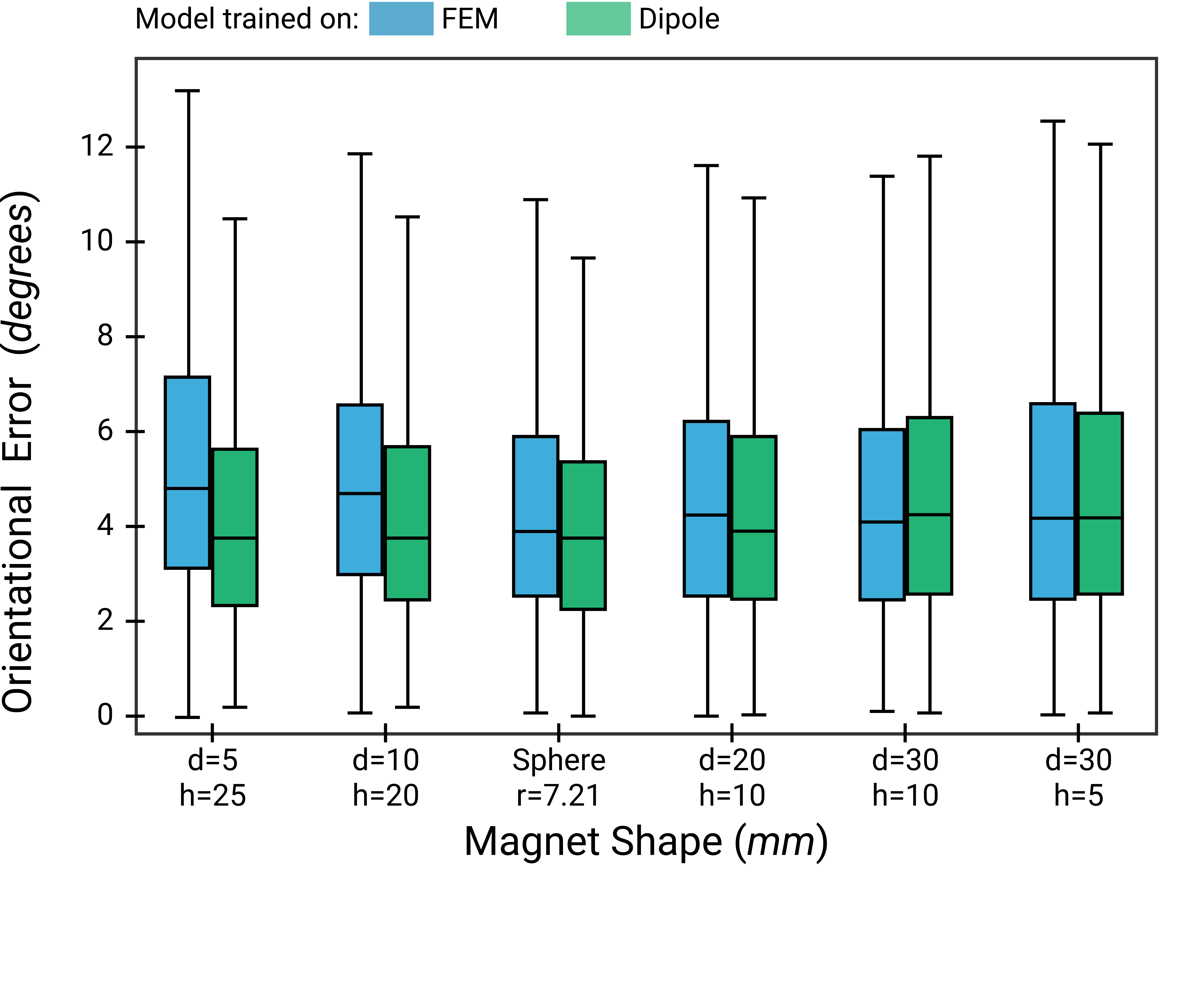}
        \caption{Angular errors}\label{fig:dipole_vs_mlp_nn_angle}
    \end{subfigure}
    \caption{Comparison of a) positional errors and b) angular errors for MLP neural networks when trained with FEM- or dipole model-generated datasets.}
\end{figure*}

\figref{fig:dipole_vs_mlp_nn} presents a comparison of the accuracy between MLPs trained on two synthetic datasets: one created from FEM simulations, as explained in \secref{sec:fem_sim}, and the other based on the magnetic dipole approximation (\eqref{eq:dipole}). Additionally, \figref{fig:dipole_vs_mlp_nn_angle} illustrates the angular error measurements. Both FEM and dipole trained methods perform well in terms of the rotational error (c.f. \figref{fig:evo_angular_error_perturb}). Overall this is an indication that our FEM-based approach generalizes well across all shapes.

To ensure the validity of our data analysis, we first conducted a Shapiro-Wilk test to confirm the normal distribution of our data. Subsequently, we applied a Student's T-Test to identify significant differences between MLPs trained with dipole- and FEM-generated datasets.

The results clearly demonstrate that our FEM-based approach significantly outperforms the dipole model-based baseline in positional accuracy for cylindrical and disk magnets across all tests (with all $p<.005$), while showing no significant difference for the spherical magnet ($p=0.95$). These results are as expected, since there is no difference between the dipole model and the FEM data for a sphere. We observed greater differences in the results as the shape of the magnet distances itself from the sphere, thereby, reinforcing our hypothesis of the superiority of our method for magnets not shaped like spheres.

\section{Experimental Evaluation}
\label{sec:opt_eval}

We compare the results of our MLP tracking method to experimental ground truth data collected with OptiTrack. We not only evaluate positional and orientation accuracy, but also evaluate the computational performance on a lightweight portable computer; enabling application such as prosthetics. We experiment with different numbers of Hall sensors in our system (4, 8, 12, or 16), to evaluate the computational speed versus accuracy trade-off. 

\subsection{Experimental Setup}
We equipped the permanent magnet with OptiTrack markers for precise tracking, illustrated in the inset of \figref{fig:sensor_loc}. The setup also features ten OptiTrack cameras, and the cylindrical magnets utilized are identical to those described in Section \secref{sec:synShapes} for simulation purposes.

In our experiments, we manipulate the magnet above the sensor grid, allowing for free movement and tilting, while ensuring optimal visibility of the optical markers to the cameras. This movement is executed at a pace similar to joystick gaming, providing a dynamic test environment. To address any discrepancies in timing between the magnetic and optical tracking systems, we implement a calibration process that involves adjusting a time-offset variable within the magnetic signal data. This adjustment aims to minimize tracking errors. Furthermore, we utilize Piecewise Cubic Hermite Interpolating Polynomial (PCHIP) interpolation  \cite{pchip}, to reconcile any discrepancies in sampling frequencies between the two systems. This approach ensures the synchronization and accuracy of our tracking data across both modalities.

\subsection{Results}
\begin{figure}[t]
    \centering
        \includegraphics[width=0.7\columnwidth]{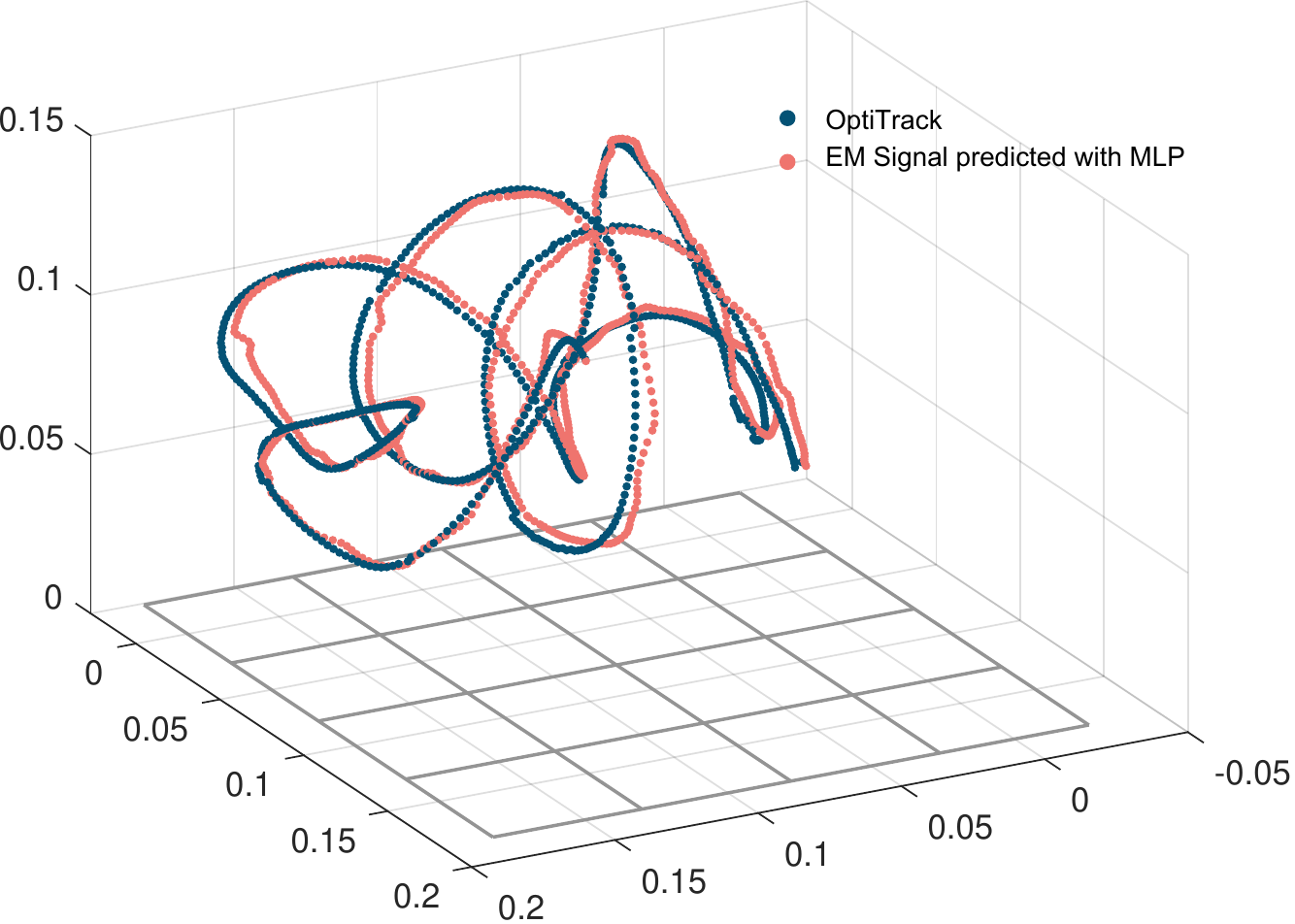}
        \caption{Comparison of trajectories obtained by OptiTrack and MLP. The difference is generally smaller than 4mm}
      \label{fig:3d_traj}
\end{figure}

\paragraph{Accuracy}
\figref{fig:3d_traj} presents a comparison of a single trajectory as tracked by both our MLP method and the OptiTrack system, aligned within the same time frame. This specific trajectory involves a magnet with dimensions of $10$mm in diameter and $20$mm in height. Throughout the tracking period, the deviation between the two trajectories remained below the $4$mm threshold for more than half of the duration, with the largest discrepancies occurring during rapid maneuvers but not exceeding $10$mm. It's important to note that the discontinuities observed in the OptiTrack trajectories can be attributed to partial occlusion of the infrared markers.

\begin{figure}[t]
    \centering
        \begin{subfigure}[b]{0.49\columnwidth}
        \includegraphics[width=\textwidth]{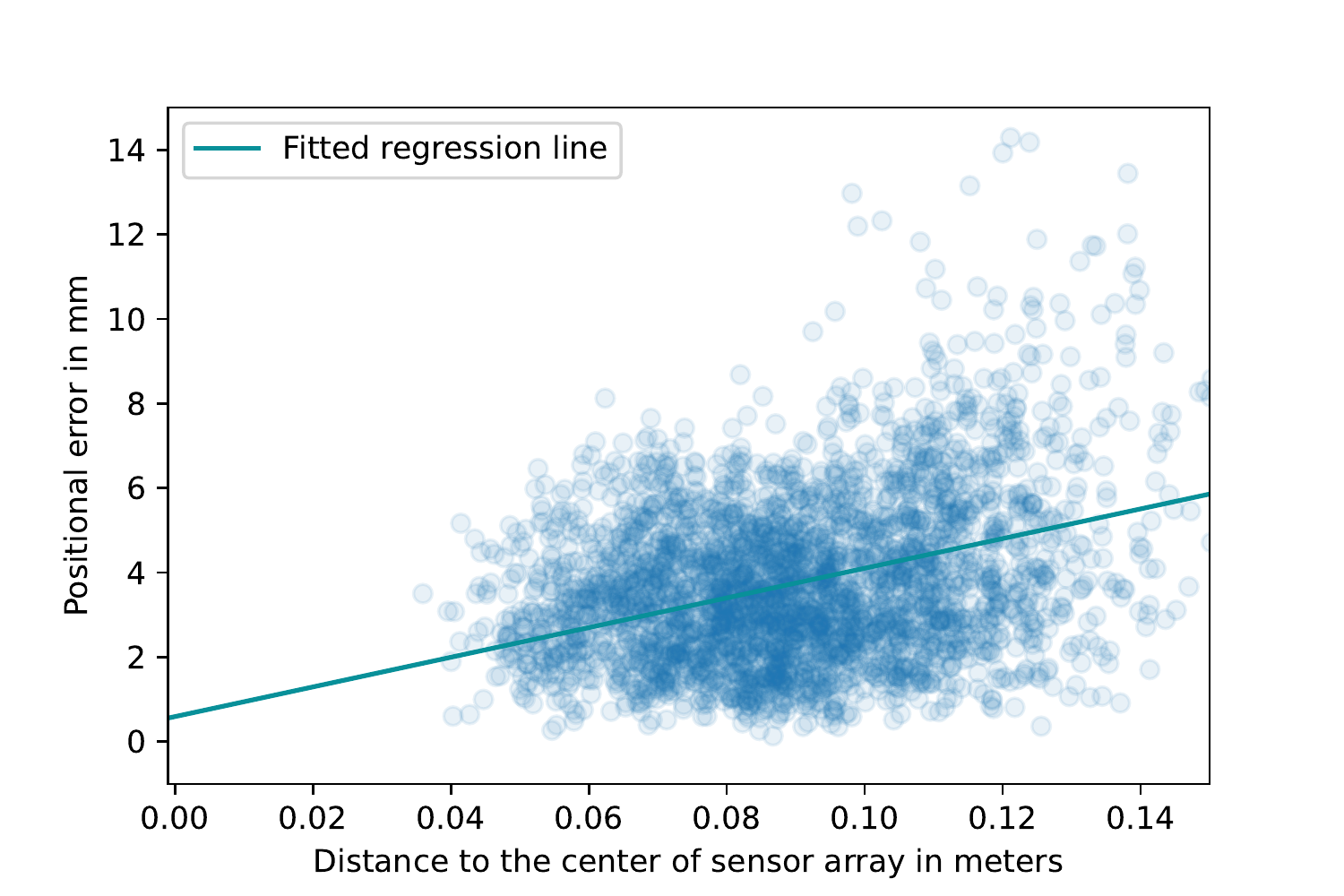}
        \caption{}
        \end{subfigure}
        \hfill
        \begin{subfigure}[b]{0.49\columnwidth}
        \includegraphics[width=\textwidth]{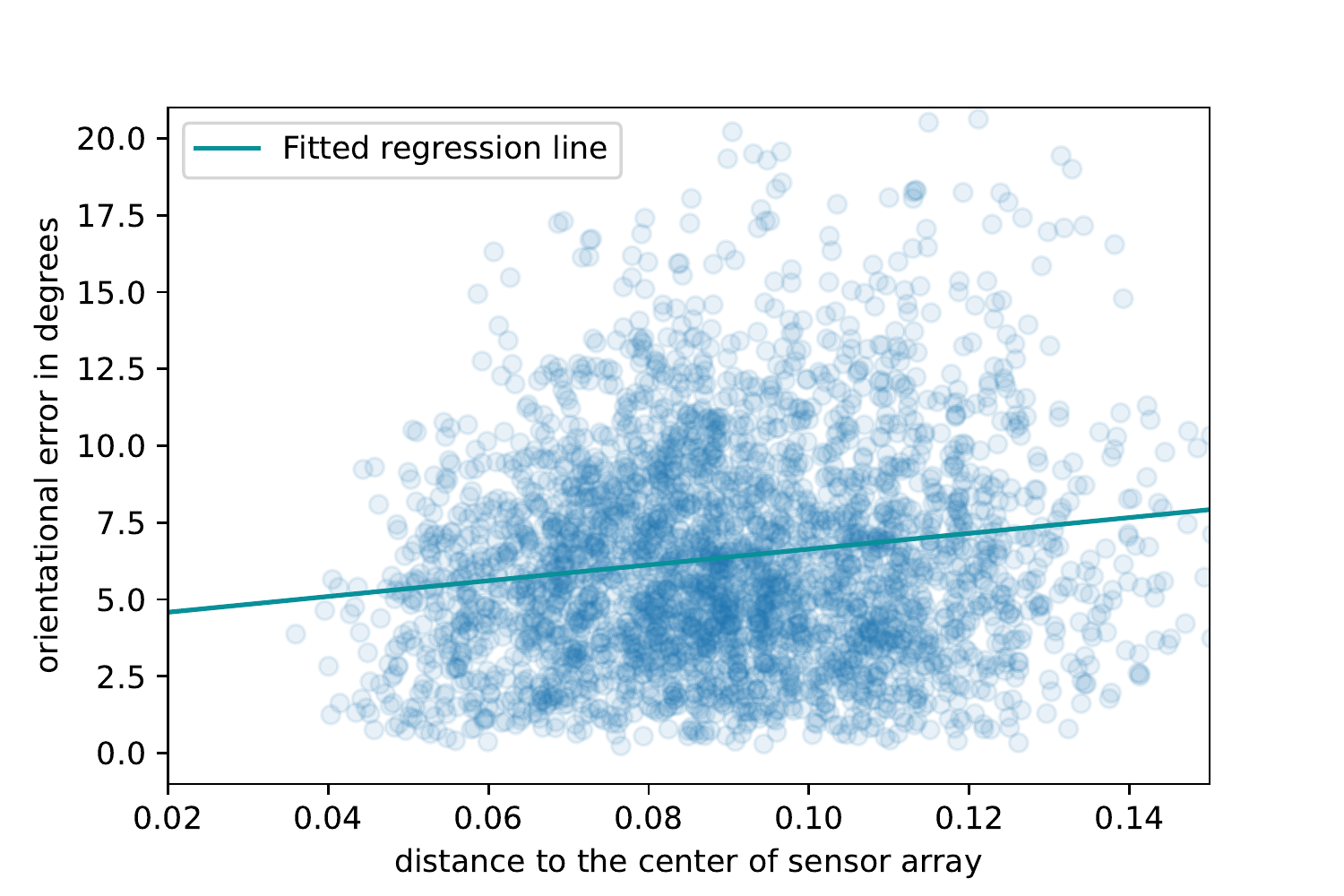}
        \caption{}
        \end{subfigure}
    \caption[]{
    \textup{(a)} Positional tracking errors versus distance between the magnet position and the center of the sensor array.
    \textup{(b)} Orientational tracking errors versus distance between the magnet position and the center of the sensor array. 
      \label{fig:error_vs_dist}}
\end{figure}

We show (\figref{fig:error_vs_dist}) the  positional error as a function of distance from the center of the sensor array, demonstrating the robustness of our tracking method even at extended ranges. Furthermore we see that the rotation error is stable for all distances, once again highlighting the robustness of our method. 

\begin{figure}[t]
    \centering
    \begin{subfigure}[b]{0.5\columnwidth}
    \includegraphics[width=\linewidth]{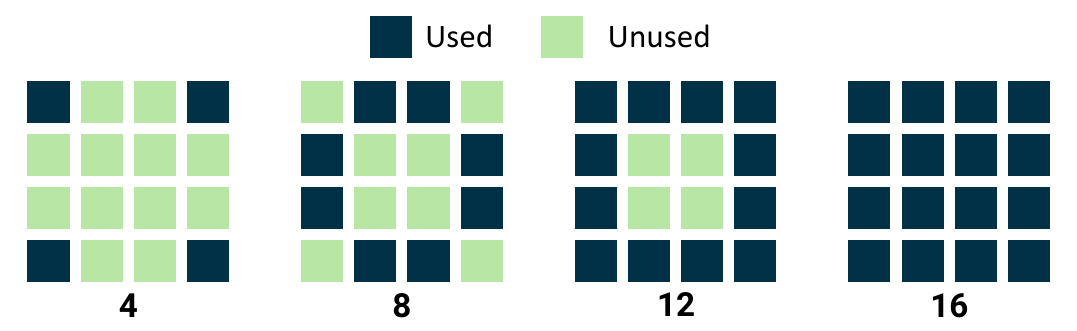}
    \caption{}
    \end{subfigure}
    
    \begin{subfigure}[b]{0.8\columnwidth}
    \includegraphics[width=\linewidth]{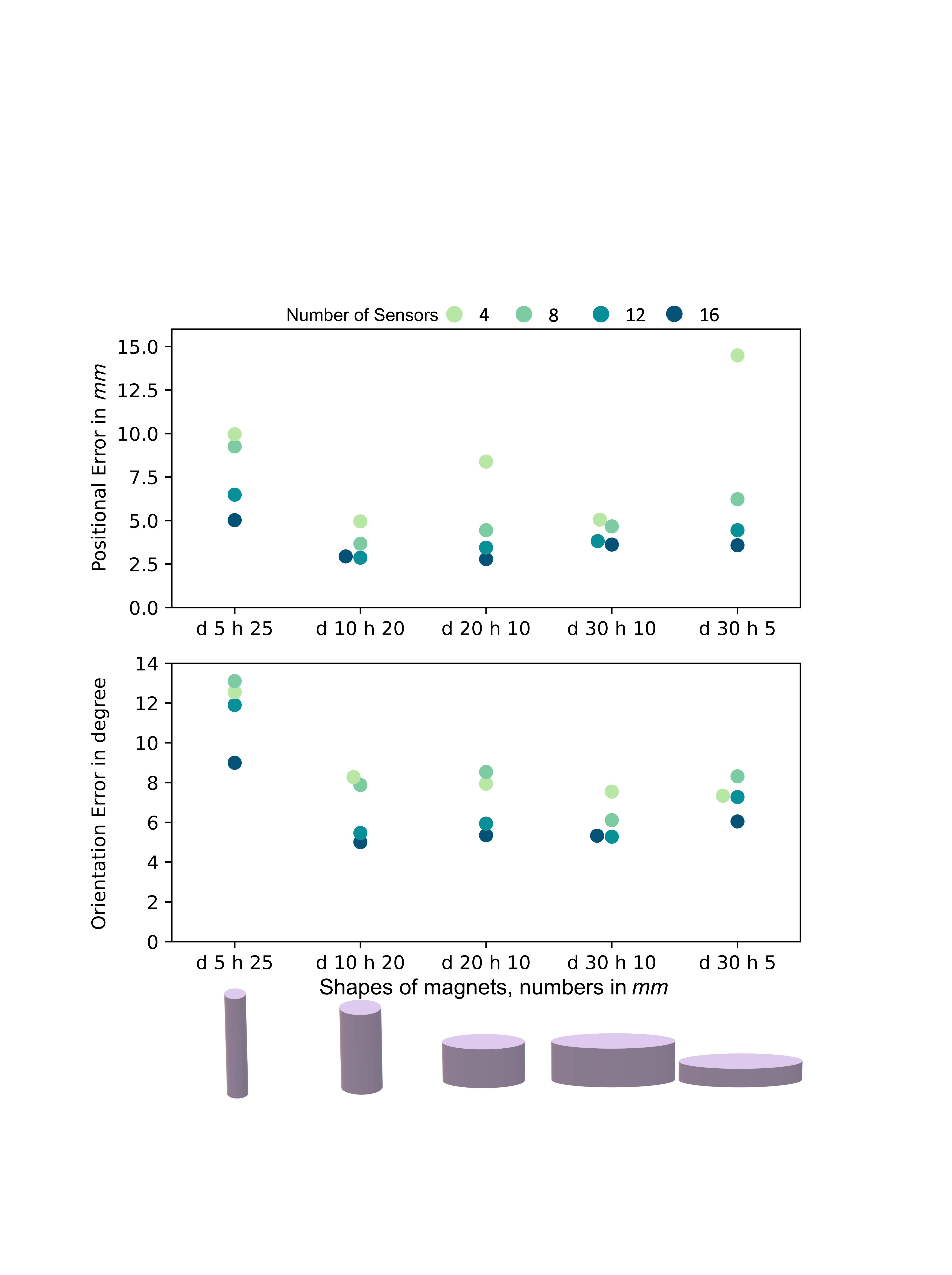}
    \caption{}
    \end{subfigure}
    \caption[]{Experimental positional and orientation errors obtained with MLP tracking, for different numbers of input sensor and magnet shapes.
      \label{fig:mlp_errors}}
\end{figure}

In \figref{fig:mlp_errors}, we detail the tracking performance statistics for five different cylindrical magnets and number of sensors. Consistent with expectations, the use of a greater number of sensors results in reduced tracking errors for all magnets tested. With the entire array of $16$ sensors activated, the average errors in both position and orientation typically fall below $4$mm and $8^\circ$ respectively, with the exception of the pole-shaped magnet (with dimensions of $5$mm in diameter and $25$mm in height).


\paragraph{Computational Time}
Finally, we evaluated the responsiveness of our tracking method by comparing the performance of inferences executed on the CPU versus the GPU of the Jetson Nano. It's crucial to highlight that sensor data collection was consistently handled by the CPU through the I2C communication protocol, taking an average time of 1.75 ms for each sensor read.

When focusing on the inference aspect, running the process on the CPU took approximately 28 ms, and this duration remained fairly consistent regardless of increases in the input feature size (i.e., number of sensors). On the other hand, executing the inference on the GPU showed a variation in processing times, starting at 10 ms with a setup of 4 sensors and extending up to 15ms when using the full array of 16 sensors. This variation underscores the GPU's ability to efficiently manage larger datasets, albeit with a modest increase in processing time as the number of sensors—and consequently, the dimensionality of the input data—grows.


\section{Application Example}
\begin{figure}
    \centering
    \includegraphics[width=0.7\columnwidth]{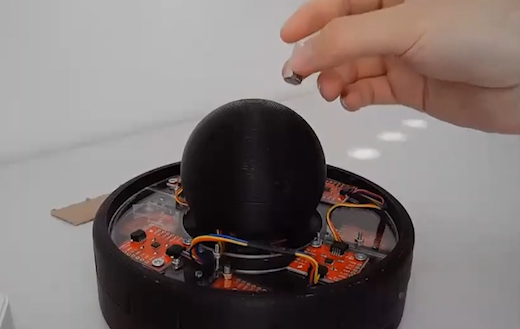}
    \caption{A user using a permanent magnet as input device for a haptic system \cite{Omni}. It is easy to imagine this magnet being embedded in an otherwise passive tool. The sphere on the device houses an haptic electromagnetic actuator.}
    \label{fig:demo}
\end{figure}

To show applicability of our approach we implemented it on a novel haptic in- and output device (\figref{fig:demo})\cite{Omni}. This implementation consists of eighth Hall sensor configured in a circle on two different level-planes. The positions and angles are tracked directly on the Jetson Nano. The latency is below 40ms and the tracking frequency reaches up to 83Hz, which is more than sufficient for most interactive applications. We refer to the supplementary material for a video of the demonstration.   

\section{Discussion}
In \secref{sec:synResults}, we identified a known limitation of iterative methods: tracking accuracy heavily depends on the initial estimation, particularly the magnet's initial orientation. The optimization often becomes trapped in a local minimum if the initial orientation estimate significantly deviates from the actual value.

However, as \figref{fig:evo_angle_perturb} demonstrates, with a reasonably accurate initial orientation (mismatch less than $45^\circ$), the convergence is less susceptible to other perturbations. The results after 50 iteration steps, as seen in \tabref{tab:sim_opt_stats}, reveal that the third quartile errors are as significant as the initialization errors. This finding suggests that the optimization method might not reach the global minimum even after 50 epochs.

In contrast, tracking with neural networks is independent of initialization. \tabref{tab:sim_opt_stats} shows that the MLP can surpass the optimization-based method in all cases except those with the most accurate initial estimations and the maximum number of iterations. Furthermore, both the third quartile positional and orientation errors are consistently within acceptable limits when using MLP, underscoring its stability.

Remarkably, we were able to directly apply the models trained with simulations in experimental tests without adjusting any hyperparameters. However, we observe a simulation-to-real gap where performance declines when training on synthetic data and evaluating on real data. This is likely due to sensor and background noise, or because our real-world magnetic markers are not perfect. For future work, we consider incorporating background and sensor noise during neural network training, akin to approaches used in other magnetic tracking systems \cite{Shao2019, Li2021}. Furthermore, we can fine-tune on real-world data. Finally, future work could include investigating additional inputs to the neural network that describe the properties of the realworld magnet (such as a magnetization scalar). These adaptations will likely diminish the simulation-to-reality gap, enhancing the accuracy of the MLP method.

In \secref{sec:synShapes}, we examined the impact of utilizing training datasets generated either through FEM or the magnetic dipole model. As anticipated, models trained with the dipole method showed performance more closely aligned with FEM-trained models when the magnet's shape was closer to or exactly a sphere. For cylindrical magnets, models trained with FEM simulations improved tracking accuracy by 0.2 mm to 1.2 mm. However, we also noted that the tracking performance of the MLP degraded as the magnet shapes deviated further from a spherical form (see \figref{fig:mlp_errors}). This may be due to demagnetizing factors not included in this study.

We observed that the total time for one-shot inference in MLP was comparable to each iteration of the many required by iterative methods. Notably, the MLP can be activated sporadically on demand, without the need to continuously track and lock the target to ensure correct convergence within a few iteration steps. We also demonstrated the feasibility of implementing an MLP tracking algorithm on a portable, energy-constrained device, such as the Jetson Nano. We found that the inference time using the GPU was about the same as reading 8 sensors via the I2C protocol on the CPU. Thus, the sensor reading process currently limits the refresh rate of our prototype. Alternative protocols like SPI could potentially alleviate this bottleneck.

A limitation of data-driven methods is the requirement to retrain the neural network for each new condition, such as different numbers and placements of sensors or changes in the magnet's shape. Although our training process, including data generation, takes only about 1 hour, this requirement could hinder applications that necessitate online optimization of sensor location. The training process could be facilitated with certain trade-off on data variety by storing and reusing generated data points in each epoch of training, rather than generating new data. Other promising directions for future research include employing neural networks to track multiple magnets, using neural network predictions to initialize optimization-based methods, and exploring the use of recurrent neural networks to enhance temporal consistency.
\section{Conclusion}
In this paper, we demonstrated the accuracy and efficiency of using neural networks to predict the location and orientation of magnets directly. We combined 2D FEM-simulated data with a coordinate transformation algorithm to generate synthetic training data on demand for any type of axis-symmetric magnet. The tracking performance of neural networks was stable and did not experience the convergence issues often seen in optimization-based tracking methods. Our experiments also showed that it is feasible to move the tracking algorithms to energy-restricted devices, thereby enabling portable interactive magnetic applications. 

\bibliographystyle{plain}
\bibliography{bibliography.bib}
\end{document}